%% file: main.tex
\PassOptionsToPackage{table,xcdraw}{xcolor}
\documentclass{article}
\usepackage{preprint,times}

\input{math_commands.tex}

\usepackage{url}
\usepackage{wrapfig}

\usepackage{graphicx}
\usepackage{xcolor}
\usepackage{colortbl}
\usepackage{booktabs}
\usepackage{multirow}
\usepackage{makecell}
\usepackage{array}
\usepackage{amssymb}
\usepackage{placeins}
\usepackage{caption}
\usepackage{float}
\usepackage{xspace}
\usepackage{fontawesome5}
\usepackage[skins,breakable]{tcolorbox}
\usepackage[colorlinks=true,linkcolor=linkblue,citecolor=linkblue,urlcolor=linkblue]{hyperref}

\definecolor{linkblue}{HTML}{1E3A8A}
\definecolor{BestBg}{HTML}{D4EDDA}
\definecolor{HeaderBg}{HTML}{F8F9FA}

\newcommand{\best}[1]{\cellcolor{BestBg}\textbf{#1}}

\newcommand{\sbayes}{\mbox{\textup{\textsc{Shape-Bayes}}}\xspace}

\newtcolorbox{keyresult}[1]{
  colback=blue!3!white, colframe=blue!50!black,
  boxrule=1pt, leftrule=4pt, arc=3pt, boxsep=0pt,
  top=1.5mm, bottom=1.5mm, left=2mm, right=2mm,
  before skip=6pt, after skip=8pt,
  fonttitle=\bfseries, coltitle=blue!50!black, colbacktitle=blue!3!white,
  attach title to upper={\ --\ }, title={#1}}

\title{Shape-Bayes: Bayesian Inference of\\ Structured Shapes under Visual Ambiguity}
\hypersetup{
  pdftitle={Shape-Bayes: Bayesian Inference of Structured Shapes under Visual Ambiguity},
  pdfauthor={Mani Kumar Tellamekala, Tosh Brown, Michel Valstar}}

\author{
  Mani Kumar Tellamekala$^{{\scalebox{1.0}{$\clubsuit$}}, 1}$  \quad \quad Tosh Brown$^2$ \quad Michel Valstar$^{2,3}$ \\
  $^1$Amsterdam, The Netherlands \quad \quad \enskip \hspace{0.05em} $^2$BlueSkeye AI, $^3$University of Nottingham, U.K. \\
  \texttt{mani.kumar.t@outlook.com} \quad \texttt{\{tosh,michel\}@blueskeye.com}
}

\begin{document}

\maketitle

\begingroup\renewcommand\thefootnote{}\footnotetext{$^{{\scalebox{1.0}{$\clubsuit$}}}$ Work independently conceived and developed by the first author during their tenure at BlueSkeye AI}\endgroup

\begin{figure}[!ht]
    \vspace{-4mm}
    \centering
    \includegraphics[width=\textwidth]{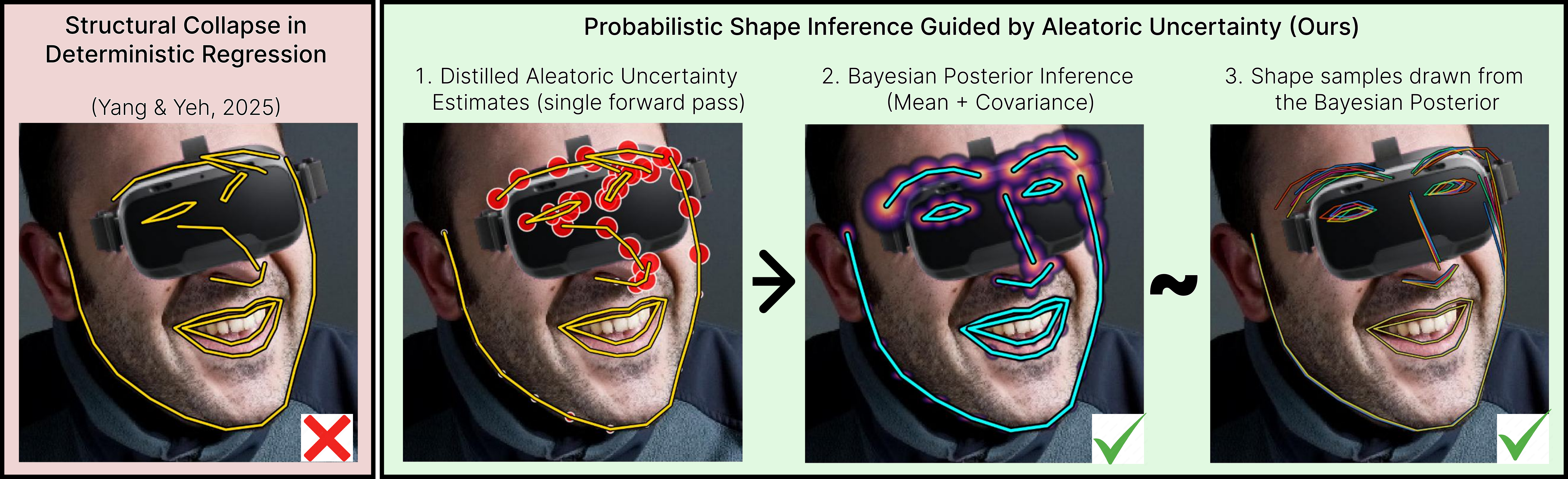}
    \caption{\textbf{Robust Bayesian shape inference under severe occlusion.} (\textbf{Left}) Deterministic regression exhibits structural collapse when forced to predict point estimates for obscured regions. (\textbf{Right}) \textbf{\sbayes} probabilistically reconciles noisy visual evidence and a latent shape prior via aleatoric uncertainty, and infers a well-calibrated posterior. This not only guarantees structural integrity but enables the sampling of multiple plausible shape hypotheses under severe ambiguity.}
    \label{fig:teaser}
    \vspace{-2mm}
\end{figure}

\begin{abstract}
\input{sections/abstract}
\end{abstract}

\input{sections/introduction}
\input{sections/method}
\input{sections/experimental_setup}
\input{sections/results}
\input{sections/related_work}
\input{sections/conclusion}

\bibliography{references}
\bibliographystyle{preprint}

\clearpage
\appendix
\raggedbottom
\input{sections/appendix}

\end{document}

%% file: math_commands.tex
\usepackage{amsmath,amsfonts,bm}

\def\eqref#1{equation~\ref{#1}}

\def\1{\bm{1}}

\DeclareMathAlphabet{\mathsfit}{\encodingdefault}{\sfdefault}{m}{sl}
\SetMathAlphabet{\mathsfit}{bold}{\encodingdefault}{\sfdefault}{bx}{n}



%% file: sections/abstract.tex
Perceiving structured shapes, such as human faces, from pixels is an inherently ambiguous task in real-world conditions. Yet, shape inference is largely posed as a deterministic regression task predicting fixed spatial coordinates. We find that deterministic regression is brittle when visual evidence is ambiguous or incomplete; under severe occlusions deterministic models exhibit structural collapse, predicting incoherent shapes or reverting to generic averages. To address this, we introduce \textbf{\sbayes}, a probabilistic framework that couples uncertainty-aware visual perception with Bayesian shape reasoning. Rather than forcing point estimates, \sbayes dynamically weights visual evidence against geometric priors to infer a structurally valid shape posterior. Demonstrated on human face shape regression, a rigorous testbed featuring complex non-rigid deformations and strict anatomical constraints, \sbayes comprises: (1) a base model predicting noisy landmarks alongside distilled aleatoric uncertainties; (2) a lightweight Transformer encoding these observations into an adaptive prior over a PCA shape manifold; and (3) a differentiable Bayesian solver computing closed-form posteriors by balancing the noisy predictions against this prior. By guaranteeing complete structural integrity, \sbayes achieves an absolute improvement of up to $\sim$34\% IDR over state-of-the-art deterministic models. Simultaneously, it yields highly calibrated uncertainty bounds and reduces relative error by up to 12.5\%, establishing a new state-of-the-art for robust 2D face shape regression under severe occlusion. The project page is at \url{https://shape-bayes.github.io} 

%% file: sections/introduction.tex
\section{Introduction}

\begin{figure}[t]
    \centering
    \includegraphics[width=0.8\textwidth]{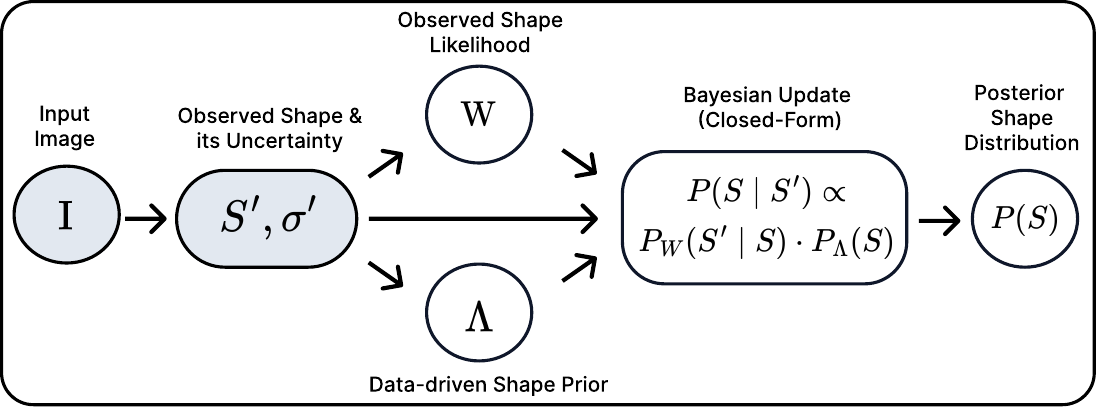}
    \caption{\textbf{Schematic overview of \sbayes.} Given an initial noisy shape estimate, \sbayes dynamically balances point-wise observation likelihoods ($W$) against a data-driven structural prior ($\Lambda$) via a closed-form Bayesian update to recover a calibrated posterior distribution $P(S)$. Although demonstrated only on face shape regression here, \sbayes is topology-agnostic and applicable to any other structured shape inference.}
    \label{fig:intro_schematic}
\end{figure}

Perceiving complex, articulated shapes is an inherently ambiguous task, particularly when visual features are occluded. Cognitive science has long posited that human vision resolves visual ambiguity through Bayesian inference~\citep{kersten2004object, mamassian2006bayesian}: the visual system continuously reconciles incomplete, noisy bottom-up observations with robust top-down structural priors. Classic computer vision formalized these probabilistic foundations for shape inference~\citep{mio2004learning, feldman2006bayesian}, principles that have recently seen a resurgence in 3D generative modeling~\citep{xu2024bayesian}. In contrast, modern shape regression has largely shifted away from these Bayesian origins in favor of deterministic point-estimate paradigms. To investigate the key vulnerabilities of deterministic regression, the human face serves as a rigorous and challenging testbed. As a ubiquitous geometric structure capable of extreme rigid and non-rigid deformations, the face provides an ideal environment to study structured shape inference under severe ambiguity.

Deep deterministic regression models have driven immense progress in structural alignment tasks, achieving remarkable accuracy on clean, unoccluded targets \citep{jin2021pixel, zhou2023star, yang2025heatmap}. However, we find that deterministic models exhibit structural collapse under severe occlusion; when forced to predict point estimates for a hidden region lacking meaningful visual evidence, they predict structurally invalid shapes. As Figure~\ref{fig:teaser} shows, they output either topologically invalid geometries or regress to a generic average shape. While recent specialized architectures attempt to address this by explicitly modeling occlusion probabilities or structural relations~\citep{zhu2022occlusion, zhu2022reasoning, li2024cascaded, chiang2025orformer, xiang2026occface}, they predominantly remain deterministic, predicting a single mean shape and entirely ignoring the true distribution of probable shapes. Although some works have begun exploring uncertainty in facial alignment~\citep{kumar2020luvli, xia2023robust}, a principled framework to infer a structurally valid shape distribution under severe occlusion remains a critical missing piece.

We introduce \textbf{\sbayes}, a Bayesian shape inference framework guided by aleatoric (data) uncertainty, as illustrated in Figure~\ref{fig:intro_schematic}. The core intuition is simple: trust visual evidence where high-confidence observations exist, and rely on anatomical priors where visual evidence is ambiguous. To arbitrate between these two, we estimate distilled aleatoric uncertainty from a teacher model. This localized uncertainty acts as a spatial gate, signaling to the Bayesian update exactly which predicted coordinates are reliable. By fusing this uncertain observation with a data-driven geometric prior, \sbayes infers a posterior distribution of coherent shapes rather than a brittle point estimate.

We evaluate \sbayes on clean and severely occluded versions of 300W~\citep{300w}, COFW~\citep{cofw}, and WFLW~\citep{wflw}. Across six diverse SOTA base models, it systematically reduces occluded landmark error (decreasing $\text{NME}_{\text{occ}}$ by $12.4\%$ for HR-noSA~\citep{yang2025heatmap}) and absolute failure rates (reducing $\text{FR}_{10}$ by $13.7\%$ for PiPNet~\citep{jin2021pixel}). Crucially, \sbayes prevents structural collapse. While deterministic models degrade under severe occlusion (e.g., In-Distribution Rate (IDR) falling to $66.6\%$ for HR-noSA), our prior-constrained Bayesian update guarantees $100\%$ IDR. Furthermore, it yields well-calibrated shape posteriors, lowering Expected Calibration Error (e.g., ECE from $7.56 \rightarrow 1.97$ for HR-noSA).

Our key contributions are:
\begin{itemize}
    \item \textbf{\sbayes Framework:} A principled Bayesian shape inference framework that leverages aleatoric uncertainty to infer full shape posteriors under visual ambiguity.
    \item \textbf{Aleatoric Uncertainty Distillation:} An architecture-agnostic distillation scheme that extracts predictive variance via spatial and photometric augmentations to localize occlusion.
    \item \textbf{Differentiable Bayesian Shape Solver:} A closed-form, differentiable solver that dynamically balances localized observation noise against a Transformer-encoded shape prior.
    \item \textbf{SOTA Robustness and Accuracy in Face Shape Regression:} Guarantees absolute structural integrity, improves shape accuracy across existing SOTA baselines and three challenging in-the-wild benchmarks, and yields well-calibrated posterior shape distributions.
\end{itemize}

%% file: sections/method.tex
\section{Shape-Bayes}
\label{sec:method}

\begin{figure}[t]
    \centering
    \includegraphics[width=\textwidth]{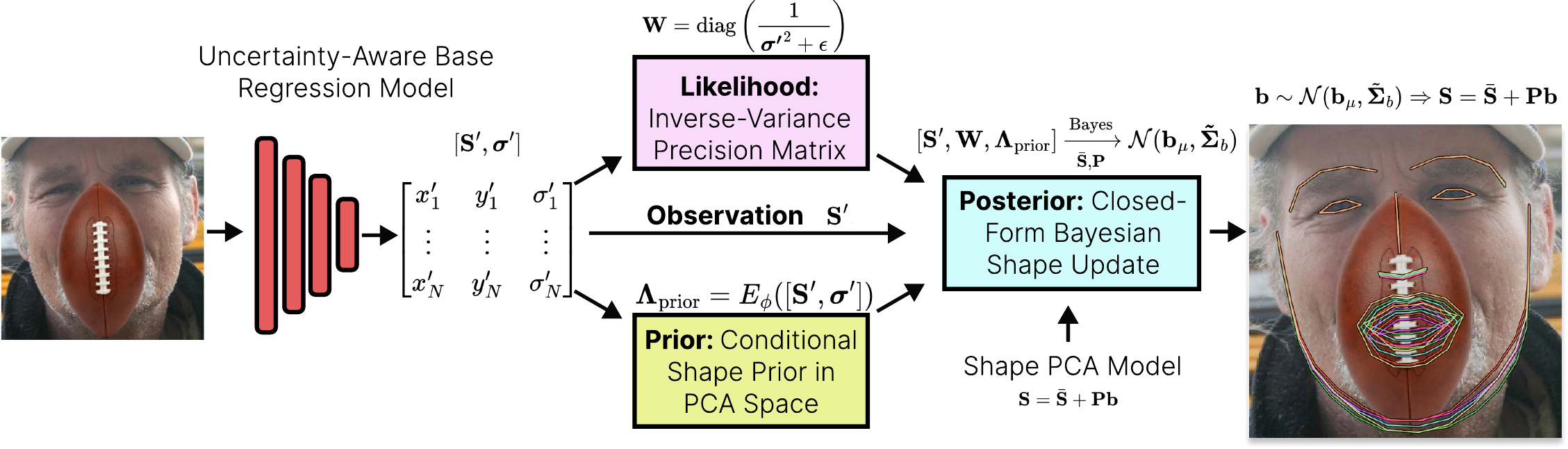}
    \caption{\textbf{\sbayes inference flow.} (1) A base regression model predicts landmarks $\mathbf{S}'$ and uncertainties $\boldsymbol{\sigma}'$, defining the likelihood precision $\mathbf{W}$. (2) A Transformer maps these observations to an adaptive prior precision $\mathbf{\Lambda}_{\text{prior}}$ in PCA space. (3) A closed-form Bayesian update computes the shape posterior $\mathcal{N}(\mathbf{b}_\mu, \tilde{\mathbf{\Sigma}}_b)$, yielding probable shapes via sampling and projection ($\mathbf{S} = \bar{\mathbf{S}} + \mathbf{P}\mathbf{b}$).}
    \label{fig:method}
\end{figure}
\begin{wrapfigure}[29]{l}{0.52\textwidth}
    \centering
    \includegraphics[width=\linewidth]{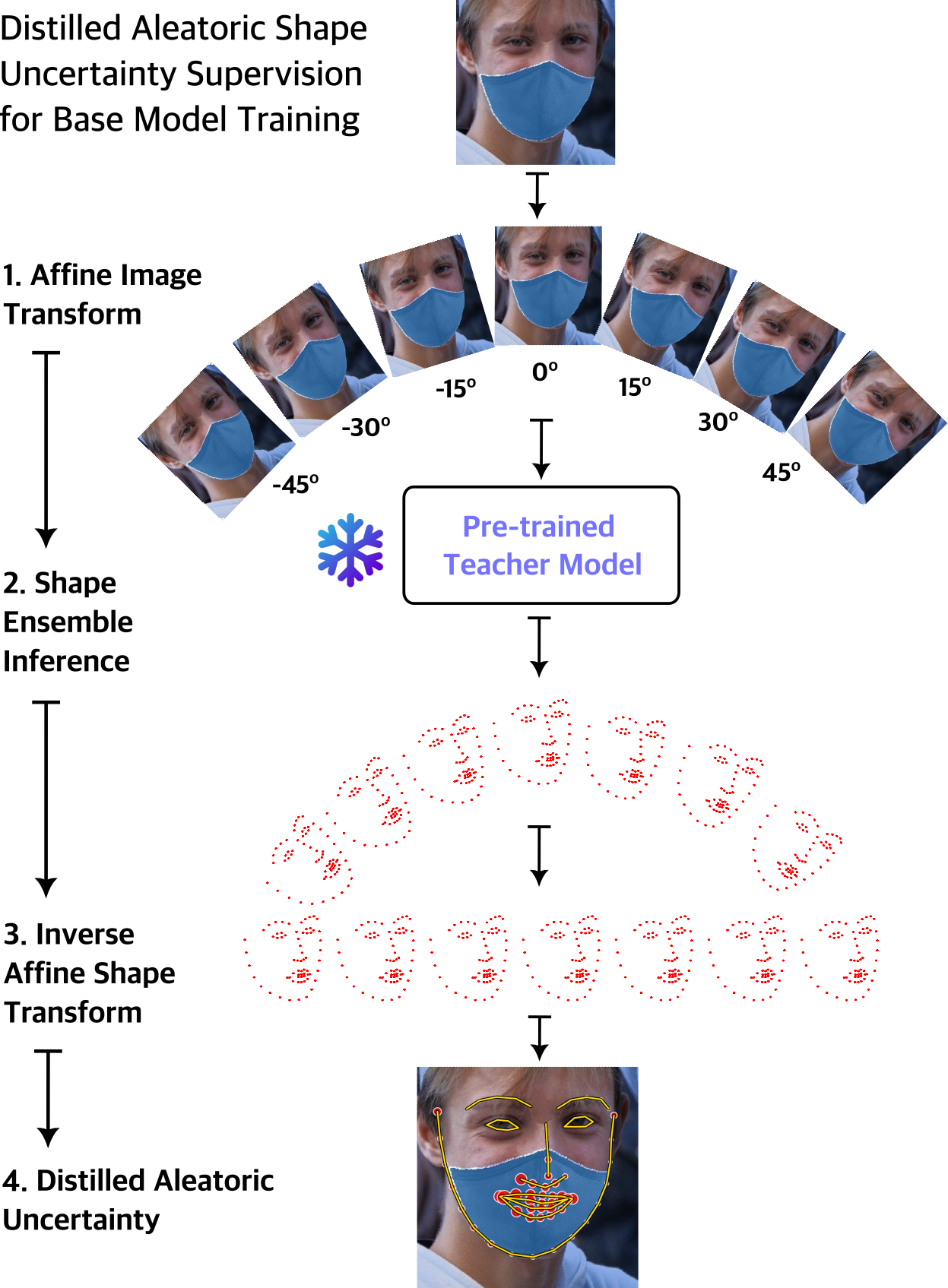}
    \caption{\textbf{Distilled Uncertainty Supervision.} Teacher predictions across affine-augmented views of an input are inverse-transformed; their empirical variance provides aleatoric targets for the base model.}
    \label{fig:uncert_distill}
\end{wrapfigure}
Deterministic shape regressors (e.g., \citet{yang2025heatmap}) exhibit structural collapse under severe occlusion (see Figure~\ref{fig:teaser}), predicting structurally invalid geometries when visual evidence is obscured. To move beyond these brittle point estimates, we introduce \sbayes, a probabilistic framework that infers a structurally valid posterior shape distribution by explicitly balancing visual observations against structural priors (Figure~\ref{fig:method}). The framework operates in three distinct steps: (1)~uncertainty-aware base regression, (2)~an observation likelihood paired with a Transformer-encoded adaptive shape prior, and (3)~exact, fully differentiable closed-form Bayesian inference.

\subsection{Uncertainty-Aware Shape Regression using a Base Model}
\label{sec:base_regression}

To effectively arbitrate between visual evidence and prior geometric knowledge, we first estimate the landmark-wise uncertainty of the observed shape. We introduce an uncertainty-aware base regression model, $\Phi_\theta$, which estimates both the initial landmark coordinates $\mathbf{S}' \in \mathbb{R}^{N \times 2}$ and their corresponding aleatoric (data) uncertainties $\boldsymbol{\sigma}' \in \mathbb{R}^{N \times 2}$.

\textbf{Supervision via Visual Feature Stability.} The model $\Phi_\theta$ is trained to estimate aleatoric uncertainty without manual annotations by exploiting the principle of \textit{visual feature stability}. A visible, salient region possesses robust local features, yielding consistent coordinate predictions even if the image is subjected to spatial or photometric stress (e.g., Gaussian blur). Conversely, an occluded region lacks discriminative evidence; a deterministic model is forced to extrapolate based on weak global context. When the image is affine transformed (e.g., rotated), such extrapolation may result in scattered predictions.

As shown in Fig.~\ref{fig:uncert_distill}, given a training image $\mathbf{I}$, we construct $M$ affine-augmented views $\{\mathcal{T}_i(\mathbf{I})\}_{i=1}^M$ (Appendix~\ref{app:base_model_training}). A frozen deterministic teacher $\Phi_T$ predicts shapes for each view. Applying the inverse transformation $\mathcal{T}_i^{-1}$ projects these coordinates back to the original image space, yielding hypotheses $\{\mathbf{S}_i\}_{i=1}^M \in \mathbb{R}^{M \times N \times 2}$. Taking the unaugmented prediction as the pseudo-ground truth target $\mathbf{S}'_{\text{target}}$, we compute the measurement uncertainty target as the ensemble's root mean square deviation: $\boldsymbol{\sigma}'_{\text{target}} = ( 1/M \sum_{i=1}^M (\mathbf{S}_i - \mathbf{S}'_{\text{target}})^2 )^{1/2}$. This establishes a heteroscedastic noise model where high local variance acts as a spatial gate, explicitly signaling regional occlusion to the Bayesian update step.

\noindent \textbf{Base Model Inference in \sbayes.} The base regression model $\Phi_\theta$ is trained using $\mathbf{S}'_{\text{target}}$ and $\boldsymbol{\sigma}'_{\text{target}}$ as supervision. This approach is highly flexible and architecture-agnostic, capable of leveraging any state-of-the-art architecture, from standard coordinate regressors to high-resolution heatmap regressors, provided the final layers can output the dual parameterization of coordinates and uncertainties. Importantly, during the \sbayes inference phase, $\Phi_\theta$ operates in a \textit{single forward pass}, predicting both the initial landmarks and their corresponding aleatoric uncertainties simultaneously. This step is crucial for inferring a reliable posterior shape distribution: by outputting landmark-wise aleatoric uncertainties, the base model acts as a probabilistic gating mechanism. It explicitly dictates to the downstream Bayesian solver which spatial regions contain reliable visual evidence to anchor the shape, and which regions must be delegated to the structural shape prior.

\subsection{Observed Shape Likelihood and Conditional Shape Prior Estimation}
\label{sec:prior}

While the predicted uncertainty $\boldsymbol{\sigma}'$ effectively flags unreliable visual evidence, correcting structural collapse requires grounding these predictions in a robust geometric prior. Although \sbayes can accommodate any differentiable shape parameterization, for our facial alignment testbed, we constrain predictions to a classic linear PCA shape manifold~\citep{cootes1995active}. This parameterizes any structurally valid shape via low-dimensional latent coefficients $\mathbf{b} \in \mathbb{R}^K$ as $\mathbf{S} = \bar{\mathbf{S}} + \mathbf{P}\mathbf{b}$, where $\bar{\mathbf{S}} \in \mathbb{R}^{2N}$ is the mean shape and $\mathbf{P} \in \mathbb{R}^{2N \times K}$ contains the principal components ($K \ll 2N$).

\noindent \textbf{Observed Shape Likelihood.} To ensure confident landmarks anchor the shape fit while discounting occluded regions, we define a Gaussian likelihood $\mathcal{N}(\mathbf{s}' \mid \bar{\mathbf{S}} + \mathbf{P}\mathbf{b}, \mathbf{W}^{-1})$ for the base model's shape prediction $\mathbf{s}' \in \mathbb{R}^{2N}$. Here, the diagonal precision matrix $\mathbf{W} = \text{diag}(1 / ({\boldsymbol{\sigma}'}^2 + \epsilon))$ explicitly acts as a landmark-wise reliability weight, gating the influence of each spatial coordinate.

\noindent \textbf{Transformer-Encoded Conditional Shape Prior.} To overcome the limitations of static linear manifolds in handling large shape deformations, a lightweight Transformer $E_\phi$ dynamically adapts the shape prior to the current occlusion pattern (Fig.~\ref{fig:method}). By processing landmark coordinates and uncertainties as $N$ tokens ($\mathbf{X} = [\mathbf{S}', \boldsymbol{\sigma}'] \in \mathbb{R}^{N \times 4}$), self-attention aggregates global geometric context, allowing confident visible regions (e.g., eyes) to deduce the structural placement of heavily occluded ones (e.g., a hidden jaw). The Transformer summarizes this context into a latent representation, which is linearly mapped to predict the diagonal elements of a prior precision matrix:
\begin{equation}
    \boldsymbol{\Lambda}_{\text{prior}} = \text{diag}(\exp(E_\phi(\mathbf{X}))) \in \mathbb{R}^{K \times K}
\end{equation}
Conceptually, $\boldsymbol{\Lambda}_{\text{prior}}$ quantifies the confidence of our structural prior \textit{before} we reconcile it with the noisy shape predictions from the base model. The exponential activation ensures strict positivity, establishing a valid, dynamically adaptive shape prior $\mathcal{N}(\mathbf{0}, \boldsymbol{\Lambda}_{\text{prior}}^{-1})$ over the latent shape coefficients.

\subsection{Differentiable Bayesian Shape Solver}
\label{sec:solver}

With the observed shape likelihood and the conditional prior established, we formulate shape inference as a Bayesian update. According to Bayes' theorem, the posterior distribution over the latent shape $p(\mathbf{b} \mid \mathbf{s}', \mathbf{X})$ is proportional to the likelihood multiplied by the prior. Because both distributions are Gaussian, the posterior is analytically exact and strictly Gaussian, $\mathcal{N}(\mathbf{b}_\mu, \tilde{\mathbf{\Sigma}}_b)$, defined by closed-form parameters:
\begin{align}
    \tilde{\mathbf{\Sigma}}_b^{-1} &= \underbrace{\mathbf{P}^T \mathbf{W} \mathbf{P}}_{\text{Observation Precision}} + \underbrace{\boldsymbol{\Lambda}_{\text{prior}}}_{\text{Prior Precision}} \label{eq:prec} \\
    \mathbf{b}_\mu &= \tilde{\mathbf{\Sigma}}_b \mathbf{P}^T \mathbf{W} (\mathbf{s}' - \bar{\mathbf{S}}) \label{eq:mean}
\end{align}

These equations formalize the arbitration between visual evidence and structural geometry. Equation~\ref{eq:prec} demonstrates that the final posterior precision ($\tilde{\mathbf{\Sigma}}_b^{-1}$) is strictly the sum of information derived from the visual observation and the structural prior. Where a region is heavily occluded, its measurement precision $\mathbf{W}$ approaches zero. Consequently, the data term $\mathbf{P}^T \mathbf{W} \mathbf{P}$ contributes zero information for that spatial dimension, and the posterior shape relies entirely on the structural coherence enforced by $\boldsymbol{\Lambda}_{\text{prior}}$. Because the manifold dimension $K$ is small (e.g., $K=70$), computing the covariance matrix $\tilde{\mathbf{\Sigma}}_b$ via inversion for Equation~\ref{eq:mean} is computationally trivial and fully differentiable.

\subsection{Training and Optimization}
\label{sec:optimization}

With the trained base regression model $\Phi_\theta$ frozen and the PCA model pre-computed, we optimize the Transformer encoder $E_\phi$ on occlusion-augmented shapes (see Appendix~\ref{app:shape_bayes_training}). The training objective anchors absolute shape coordinates while simultaneously calibrating the predicted posterior distribution. Let $\mathbf{b}_{\text{gt}}$ be the ground truth PCA coefficients. The joint optimization objective is:
\begin{equation}
    \mathcal{L} = \lambda_{\text{MSE}} \mathcal{L}_{\text{MSE}} + \lambda_{\text{NLL}} \mathcal{L}_{\text{NLL}}
\end{equation}

The first term is a standard Mean Squared Error (MSE) loss on the posterior mean, $\mathcal{L}_{\text{MSE}} = \|\mathbf{b}_\mu - \mathbf{b}_{\text{gt}}\|^2$, driving landmark localization accuracy. The second term is the Negative Log-Likelihood (NLL) of the ground truth shape under the predicted posterior:
\begin{equation}
    \mathcal{L}_{\text{NLL}} = \frac{1}{2} (\mathbf{b}_{\text{gt}} - \bar{\mathbf{b}}_\mu)^T \tilde{\mathbf{\Sigma}}_b^{-1} (\mathbf{b}_{\text{gt}} - \bar{\mathbf{b}}_\mu) - \frac{1}{2} \log\det(\tilde{\mathbf{\Sigma}}_b^{-1})
\end{equation}

Crucially, applying a stop-gradient to the mean ($\bar{\mathbf{b}}_\mu = \text{sg}[\mathbf{b}_\mu]$) during NLL loss computation decouples optimization dynamics: MSE drives coordinate prediction (the \textit{where}), while NLL strictly optimizes prior precision (the \textit{how sure}), preventing the model from artificially lowering the loss by shifting only the mean shape instead of refining its variance.

%% file: sections/experimental_setup.tex
\section{Experimental Setup}
\label{sec:experimental_setup}

We evaluate \sbayes on 2D face shape regression, a canonical geometric testbed exhibiting complex rigid and non-rigid deformations, under strict anatomical constraints. Our evaluation covers point-estimation accuracy, structural integrity, and posterior calibration under severe occlusions.

\noindent \textbf{Datasets and Dynamic Occlusion.} We evaluate on the test sets of {300W}~\citep{300w}, {COFW}~\citep{cofw}, and {WFLW}~\citep{wflw}. To test out-of-distribution robustness without altering ground-truth shapes, we apply a dynamic occlusion protocol that procedurally overlays synthetic wearable masks and a library of 80+ real-world objects onto images. This yields clearly occluded landmark indices for evaluation. More details can be found in Appendix~\ref{app:occlusion_assets}.

\noindent \textbf{Base Models and Aleatoric Uncertainty Distillation.} To demonstrate its versatility, we integrate \sbayes into six diverse state-of-the-art base models spanning heatmap, coordinate, and transformer-based paradigms: HR-noSA~\citep{yang2025heatmap}, LUVLi~\citep{kumar2020luvli}, DSLPT~\citep{xia2023robust}, ORFormer~\citep{chiang2025orformer}, STAR~\citep{zhou2023star}, and PiPNet~\citep{jin2021pixel}. These base models are fine-tuned to jointly predict coordinates $\mathbf{S}'$ and distilled aleatoric uncertainty $\boldsymbol{\sigma}'$ from their corresponding pre-trained teacher models (see Appendix~\ref{app:base_model_training}).

\noindent \textbf{Implementation.} \sbayes comprises two key components: (1) a \textit{Latent PCA Manifold} (a rank-$K$ subspace capturing $>99\%$ of structural variance), and (2) a \textit{Prior Transformer} (a lightweight 5-layer encoder, $d_{\text{model}}=64$) that maps noisy observations $[\mathbf{S}', \boldsymbol{\sigma}']$ to prior precision parameters $\boldsymbol{\Lambda}_{\text{prior}}$. To ground the prior in empirical noise, we train this Transformer using corrupted triplets $(\mathbf{S}_{\text{gt}}, \mathbf{S}', \boldsymbol{\sigma}')$ sampled from a base model (Appendix~\ref{app:shape_bayes_training}). Crucially, \sbayes adds negligible computational overhead ($<0.2\%$ FLOPs relative to HR-noSA~\citep{yang2025heatmap}).

\noindent \textbf{Evaluation Metrics.} \textit{(1) Localization Accuracy:} We report Normalized Mean Error (global $\text{NME}_{\text{all}}$, occluded $\text{NME}_{\text{occ}}$), minimum NME ($\text{minNME}$) across 100 posterior samples (Appendix~\ref{app:diverse_sampling}), Failure Rate ($\text{FR}_{10}$) and $\text{AUC}_{10}$ (using the standard 10\% NME threshold). \textit{(2) Structural Integrity:} Following \citet{lee2018simple}, we quantify structural collapse via the {In-Distribution Rate (IDR)}, the percentage of shape predictions falling within the $95\%$ confidence hyperellipsoid of the learned PCA manifold:
\begin{equation}
\text{IDR} = \frac{1}{|\mathcal{D}|} \sum_{i \in \mathcal{D}} \mathbb{I} \left( (\mathbf{b}_i - \boldsymbol{\mu})^T \boldsymbol{\Sigma}_{\text{PCA}}^{-1} (\mathbf{b}_i - \boldsymbol{\mu}) \le \chi^2_K(0.95) \right)
\end{equation}
where $\mathbf{b}_i$ is the projected latent vector, $\boldsymbol{\mu}, \boldsymbol{\Sigma}_{\text{PCA}}$ are training manifold statistics, and $\chi^2_K(0.95)$ is the chi-squared critical value. \textit{(3) Probabilistic Calibration:} We evaluate posterior uncertainty bounds using Negative Log-Likelihood ($\text{NLL}$), Expected Calibration Error ($\text{ECE}$), and Spearman rank correlation ($\rho$) between predicted landmark variance and true Euclidean error.

%% file: sections/results.tex
\section{Results and Discussion}
\label{sec:results}

\begin{figure}[t]
    \centering
    \includegraphics[width=\textwidth]{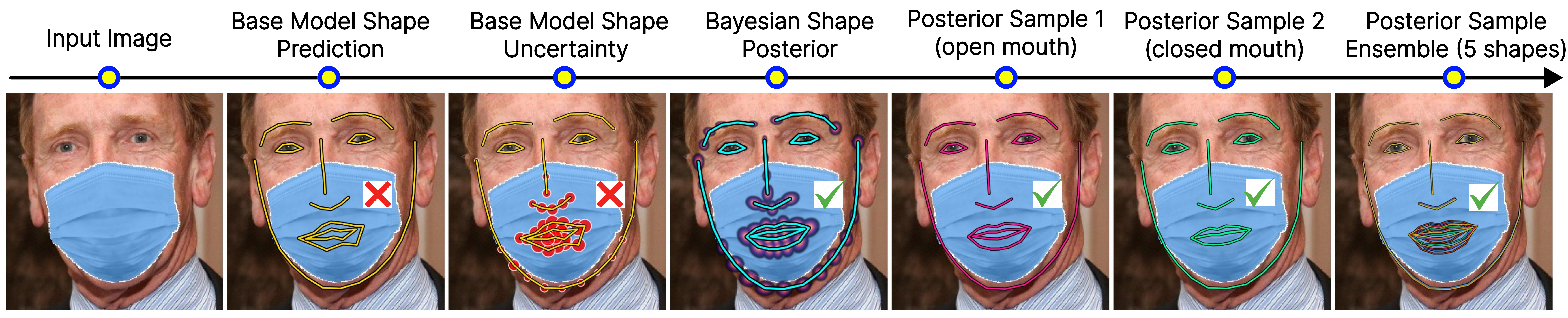}
    \caption{\textbf{Qualitative illustration of the \sbayes inference pipeline.} From left to right: input image, base model shape prediction and its aleatoric uncertainty, Bayesian posterior mean, and shapes sampled from the posterior.}
    \label{fig:bayes_shape_flow}
\end{figure}

We evaluate whether \sbayes can recover structurally valid shapes without compromising accuracy under  occlusions. Its core hypothesis is that conditioning a bottom-up noisy shape prediction and its aleatoric uncertainty on a top-down shape prior yields well-calibrated shape posteriors. We test this hypothesis along three axes: \textit{(1) structural integrity of the posterior mean, (2) shape accuracy across deterministic and uncertainty-aware base models, and (3) posterior calibration.}

\noindent \textbf{Structural Integrity Evaluation.} The inability of standard regression models to output structurally valid face shapes under severe occlusion is evident in their low In-Distribution Rate (IDR) scores (Tables~\ref{tab:softargmax_occluded} and~\ref{tab:models_all_dynamic_occlusion}). Despite achieving state-of-the-art accuracy on the unoccluded test sets, HR-noSA~\citep{yang2025heatmap} suffers severe structural collapse under heavy dynamic occlusion, dropping to an IDR of just 66.62\% on 300W. Similarly, while native uncertainty-aware baselines like LUVLi improve structural validity (82.8\% on 300W), their reliance on independent, per-landmark variance without a global shape prior still allows predictions to fall outside the valid manifold. In stark contrast, \sbayes completely resolves this vulnerability, as illustrated in Figures~\ref{fig:bayes_shape_flow} and~\ref{fig:main_300w}, achieving a perfect 100\% IDR across all six evaluated base models and all three test sets.

While base models suffer structural collapse on occluded 300W and WFLW, their IDR remains deceptively high on COFW due to its sparse 29-point topology. With fewer spatial constraints than dense 68- or 98-point formulations, distorted predictions on COFW rarely breach the manifold boundaries, making IDR a much stricter metric for dense topologies. Regardless of landmark density, \sbayes guarantees a 100\% IDR by strictly conditioning all predictions on the learned PCA prior.

\begin{table}[t]
\centering
\begin{minipage}[c]{0.29\textwidth}
    \centering
    \captionsetup{justification=raggedright,singlelinecheck=false}
    \includegraphics[width=\linewidth]{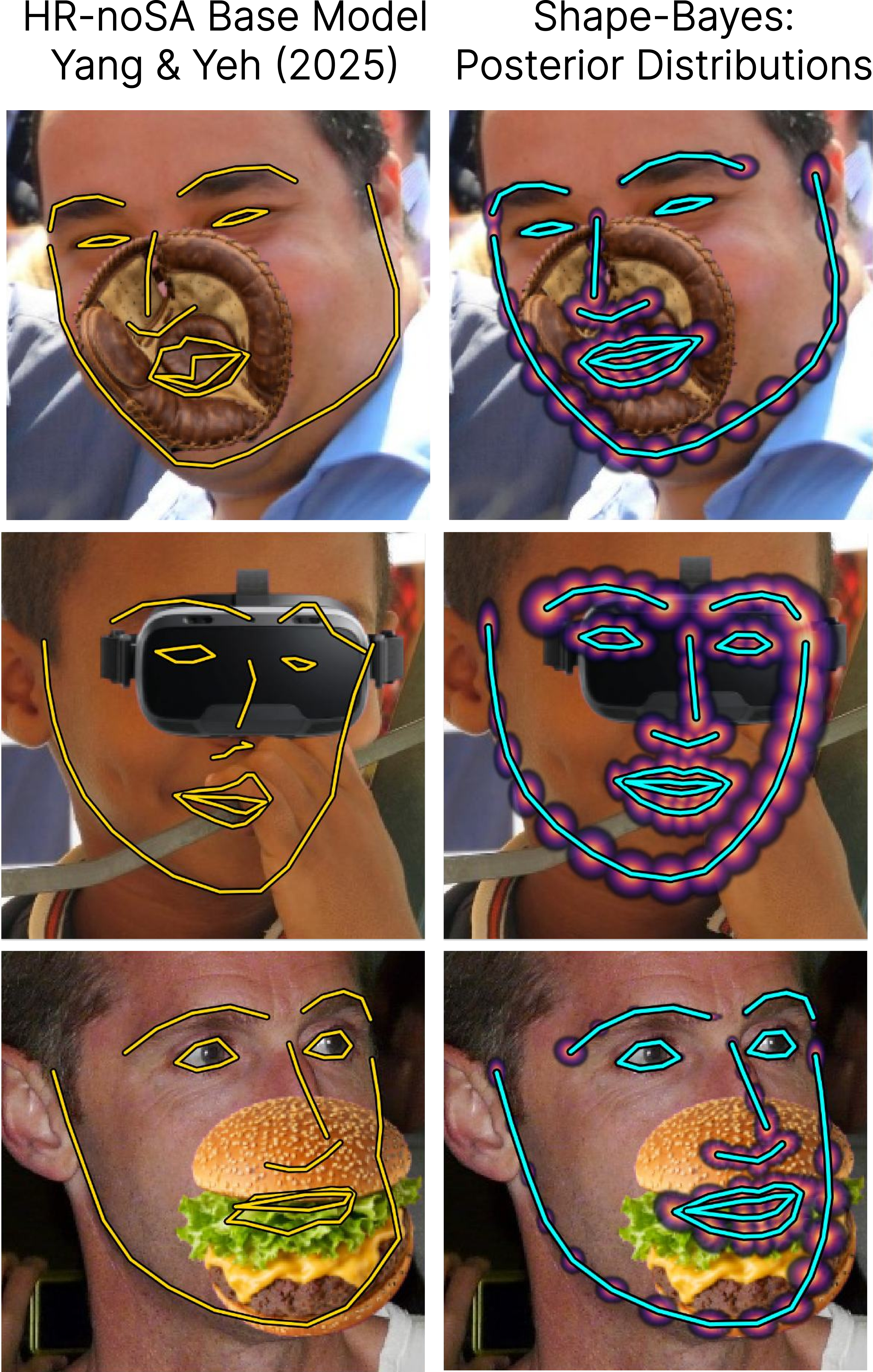}
    \captionof{figure}{\textbf{(Left)} Deterministic regression exhibits structural collapse. \textbf{(Right)} \sbayes infers robust, structurally valid posterior shapes.}
    \label{fig:main_300w}
\end{minipage}%
\hfill
\begin{minipage}[c]{0.68\textwidth}
\centering
\footnotesize
\renewcommand{\arraystretch}{1.3}
\setlength{\tabcolsep}{2.6pt}
\caption{\sbayes (+SB) improves accuracy and ensures structural integrity (100\% IDR) under dynamic occlusion (base model: HR-noSA~\citep{yang2025heatmap}).}
\label{tab:softargmax_occluded}
\begin{tabular}{l l c c c c c}
\toprule
\cellcolor{blue!12}\textbf{Data} &
\cellcolor{blue!12}\textbf{Cfg.} &
\cellcolor{cyan!12}\textbf{IDR\,$\uparrow$} &
\cellcolor{teal!12}\textbf{NME$_{\text{occ}}\,\downarrow$} &
\cellcolor{teal!20}\textbf{NME$_{\text{all}}$ \textit{(M/min)}\,$\downarrow$} &
\cellcolor{green!12}\textbf{FR\,$\downarrow$} &
\cellcolor{green!20}\textbf{AUC\,$\uparrow$} \\
\midrule
\cellcolor{white} & Base & 66 & 9.03 & 6.31 & 11.8\% & 41.3\% \\
\rowcolor{green!5} \multirow{-2}{*}{\cellcolor{white}\textbf{300W}}
& \textbf{+SB} & \textbf{100} & \textbf{7.91} & \textbf{5.90} / \textbf{5.49} & \textbf{9.4\%} & \textbf{44.3\%} \\
\midrule
\cellcolor{white} & Base & 81 & 12.34 & 9.12 & 25.6\% & 27.9\% \\
\rowcolor{green!5} \multirow{-2}{*}{\cellcolor{white}\textbf{WFLW}}
& \textbf{+SB} & \textbf{100} & \textbf{11.35} & \textbf{8.79} / \textbf{8.36} & \textbf{23.4\%} & \textbf{29.8\%} \\
\midrule
\cellcolor{white} & Base & 96 & 10.28 & 7.40 & 16.0\% & 33.3\% \\
\rowcolor{green!5} \multirow{-2}{*}{\cellcolor{white}\textbf{COFW}}
& \textbf{+SB} & \textbf{100} & \textbf{9.16} & \textbf{7.08} / \textbf{6.41} & \textbf{13.2\%} & \textbf{35.3\%} \\
\bottomrule
\end{tabular}

\vspace{6mm}
\setlength{\tabcolsep}{3.8pt}\renewcommand{\arraystretch}{1.3}
\caption{\textbf{Part-wise NME$_{\text{occ}}$\,$\downarrow$} on occluded 300W (base model: HR-noSA).}
\label{tab:300w_softargmax_regions_wrapped}
\begin{tabular}{l ccccccc}
\toprule
\cellcolor{blue!12}\textbf{Cfg.} & \cellcolor{blue!12}\textbf{Jaw} & \cellcolor{blue!12}\textbf{Mouth} & \cellcolor{blue!12}\textbf{L-Brow} & \cellcolor{blue!12}\textbf{R-Brow} & \cellcolor{blue!12}\textbf{Nose} & \cellcolor{blue!12}\textbf{L-Eye} & \cellcolor{blue!12}\textbf{R-Eye} \\
\midrule
Base & 11.2 & 10.2 & 7.6 & 7.0 & 6.6 & 5.8 & 5.2 \\
\rowcolor{green!5} \textbf{+SB} & \textbf{10.1} & \textbf{8.9} & \textbf{7.0} & \textbf{6.4} & \textbf{5.8} & \textbf{5.1} & \textbf{4.5} \\
\bottomrule
\end{tabular}
\end{minipage}
\end{table}

\begin{keyresult}{Key Result 1: Resolution of Structural Collapse}
\sbayes guarantees structural integrity in shape predictions by inferring posterior distributions over a global shape manifold.
\end{keyresult}

\noindent \textbf{Shape Prediction Accuracy Evaluation.} A common failure mode of PCA-like prior-driven shape regularization is over-smoothing, wherein structural constraints degrade fine-grained landmark accuracy on visible regions. \sbayes eliminates this trade-off: its well-conditioned posteriors preserve fine-grained precision on visible landmarks while consistently improving accuracy under occlusions. For HR-noSA under dynamic occlusion, \sbayes yields systematic gains across all datasets (Table~\ref{tab:softargmax_occluded}). On 300W, the posterior mean reduces relative $\text{NME}_{\text{occ}}$ by 12.4\% and relative $\text{NME}_{\text{all}}$ by 6.5\%, extending to a 13\% error reduction via 100-sample posterior draws. This boosts reliability, dropping absolute Failure Rate by 2.4\%. These gains remain consistent across all facial regions (Table~\ref{tab:300w_softargmax_regions_wrapped}). These results show that \sbayes successfully resolves the tension between structural integrity and fine-grained accuracy through its closed-form Bayesian solver, which employs predicted point-wise uncertainty as precision weights ($W_i = \sigma_i^{-2}$). High-confidence, visible landmarks act as strong anchors that tightly constrain the latent shape pose, whereas highly uncertain, occluded points are automatically down-weighted and guided by the conditional expectations of the geometric prior.

\begin{table*}[t]
\centering
\small
\renewcommand{\arraystretch}{1.15}
\setlength{\tabcolsep}{2.5pt}
\caption{\textbf{Generalization of \sbayes (+SB) across base models and datasets} ($\text{NME}_{\text{all}}$: mean / $\text{minNME}_{@100}$). Base models in \colorbox{gray!30}{gray} natively output uncertainty, requiring no training.}
\label{tab:models_all_dynamic_occlusion}
\resizebox{\textwidth}{!}{%
\begin{tabular}{ll ccccc ccccc ccccc}
\toprule
\cellcolor{blue!12} & \cellcolor{blue!12} &
\multicolumn{5}{c}{\cellcolor{cyan!12}\textbf{300W (Dynamic Occlusion)}} &
\multicolumn{5}{c}{\cellcolor{teal!12}\textbf{WFLW (Dynamic Occlusion)}} &
\multicolumn{5}{c}{\cellcolor{green!12}\textbf{COFW (Dynamic Occlusion)}} \\
\cmidrule(lr){3-7} \cmidrule(lr){8-12} \cmidrule(lr){13-17}
\cellcolor{blue!12}\textbf{Method} &
\cellcolor{blue!12}\textbf{Cfg.} &
\cellcolor{cyan!8}\textbf{IDR\,$\uparrow$} &
\cellcolor{cyan!8}\textbf{NME$_{\text{occ}}$\,$\downarrow$} &
\cellcolor{cyan!8}\textbf{NME$_{\text{all}}$\,$\downarrow$} &
\cellcolor{cyan!8}\textbf{FR\,$\downarrow$} &
\cellcolor{cyan!8}\textbf{AUC\,$\uparrow$} &
\cellcolor{teal!8}\textbf{IDR\,$\uparrow$} &
\cellcolor{teal!8}\textbf{NME$_{\text{occ}}$\,$\downarrow$} &
\cellcolor{teal!8}\textbf{NME$_{\text{all}}$\,$\downarrow$} &
\cellcolor{teal!8}\textbf{FR\,$\downarrow$} &
\cellcolor{teal!8}\textbf{AUC\,$\uparrow$} &
\cellcolor{green!8}\textbf{IDR\,$\uparrow$} &
\cellcolor{green!8}\textbf{NME$_{\text{occ}}$\,$\downarrow$} &
\cellcolor{green!8}\textbf{NME$_{\text{all}}$\,$\downarrow$} &
\cellcolor{green!8}\textbf{FR\,$\downarrow$} &
\cellcolor{green!8}\textbf{AUC\,$\uparrow$} \\
\midrule
\cellcolor{gray!15} & Base & 83 & 9.87 & 7.06 & 11.8 & 36.8 & 90 & 12.27 & 9.08 & 23.3 & 30.9 & -- & -- & -- & -- & -- \\
\rowcolor{gray!30} \cellcolor{gray!15}\multirow{-2}{*}{\textbf{LUVLi}} & \textbf{+SB} & \textbf{100} & \textbf{8.86} & \textbf{6.67}/\textbf{6.17} & \textbf{9.9} & \textbf{39.5} & \textbf{100} & \textbf{11.33} & \textbf{8.73}/\textbf{7.97} & \textbf{22.0} & \textbf{32.3} & -- & -- & -- & -- & -- \\
\midrule
\cellcolor{gray!15} & Base & -- & -- & -- & -- & -- & 98 & 12.08 & 8.81 & 25.3 & 28.3 & -- & -- & -- & -- & -- \\
\rowcolor{gray!30} \cellcolor{gray!15}\multirow{-2}{*}{\textbf{DSLPT}} & \textbf{+SB} & -- & -- & -- & -- & -- & \textbf{100} & \textbf{11.46} & \textbf{8.66}/\textbf{8.25} & \textbf{24.6} & \textbf{29.4} & -- & -- & -- & -- & -- \\
\midrule
\cellcolor{white} & Base & 78 & 10.36 & 7.07 & 14.4 & 36.5 & 84 & 10.49 & 8.24 & 21.3 & 33.0 & 100 & 9.56 & 6.61 & 10.8 & 37.1 \\
\rowcolor{green!5} \cellcolor{white}\multirow{-2}{*}{\textbf{ORFormer}} & \textbf{+SB} & \textbf{100} & \textbf{9.06} & \textbf{6.75}/\textbf{6.33} & \textbf{12.3} & \textbf{39.6} & \textbf{100} & \textbf{10.13} & \textbf{8.07}/\textbf{7.57} & \textbf{20.2} & \textbf{33.8} & \textbf{100} & \textbf{8.93} & \textbf{6.48}/\textbf{5.97} & \textbf{9.1} & \textbf{38.0} \\
\midrule
\cellcolor{white} & Base & 63 & 11.80 & 8.47 & 26.6 & 27.3 & 85 & 11.38 & 8.41 & 21.4 & 31.5 & 100 & 10.75 & 8.15 & 21.1 & 27.1 \\
\rowcolor{green!5} \cellcolor{white}\multirow{-2}{*}{\textbf{STAR}} & \textbf{+SB} & \textbf{100} & \textbf{9.70} & \textbf{7.51}/\textbf{6.87} & \textbf{18.6} & \textbf{34.1} & \textbf{100} & \textbf{10.69} & \textbf{8.18}/\textbf{7.69} & \textbf{20.0} & \textbf{32.9} & \textbf{100} & \textbf{9.53} & \textbf{7.79}/\textbf{7.08} & \textbf{18.3} & \textbf{29.1} \\
\midrule
\cellcolor{white} & Base & 75 & 14.88 & 10.43 & 41.2 & 18.3 & 92 & 11.07 & 8.40 & 21.4 & 31.3 & 100 & 10.56 & 8.50 & 19.7 & 28.8 \\
\rowcolor{green!5} \cellcolor{white}\multirow{-2}{*}{\textbf{PiPNet}} & \textbf{+SB} & \textbf{100} & \textbf{11.72} & \textbf{8.76}/\textbf{8.23} & \textbf{27.5} & \textbf{25.3} & \textbf{100} & \textbf{9.90} & \textbf{8.00}/\textbf{7.33} & \textbf{19.3} & \textbf{33.8} & \textbf{100} & \textbf{9.51} & \textbf{7.81}/\textbf{7.17} & \textbf{16.0} & \textbf{31.2} \\
\bottomrule
\end{tabular}}
\end{table*}

\noindent \textbf{Cross-Architecture Generalization.} \sbayes systematically improves all six base models (Table~\ref{tab:models_all_dynamic_occlusion}). On occluded 300W, PiPNet's relative $\text{NME}_{\text{occ}}$ drops 21.2\% (with a 13.7\% absolute FR reduction), while STAR reduces relative $\text{NME}_{\text{occ}}$ by 17.8\% and fully restores IDR ($63.0\% \to 100\%$). \emph{Crucially, architectures with native uncertainty heads (LUVLi, DSLPT; gray rows) require no distillation. By directly ingesting LUVLi's predicted uncertainty, \sbayes lowers relative $\text{NME}_{\text{occ}}$ by 10.2\% (300W) and 7.7\% (WFLW), showing its efficacy as a versatile, probabilistic extension.}

\begin{table*}[t]
\centering
\begin{minipage}[t]{0.45\textwidth}
\centering
\footnotesize
\renewcommand{\arraystretch}{1.5}
\setlength{\tabcolsep}{2pt}
\caption{\textbf{Results on original test sets} with HR-noSA~\citep{yang2025heatmap} as the base model.}
\label{tab:softargmax_original}
\begin{tabular}{ll cccc}
\toprule
\cellcolor{blue!12}\textbf{Dataset} &
\cellcolor{blue!12}\textbf{Cfg.} &
\cellcolor{cyan!12}\textbf{IDR\,$\uparrow$} &
\cellcolor{teal!12}\makecell{\textbf{NME$_{\text{all}}$\,$\downarrow$}\\[-1pt]\textit{\scriptsize (mean / min)}} &
\cellcolor{green!12}\textbf{FR\,$\downarrow$} &
\cellcolor{green!20}\textbf{AUC\,$\uparrow$} \\
\midrule
\cellcolor{white} & Base & 99 & \textbf{2.86} & \textbf{0.1} & \textbf{71.4} \\
\rowcolor{green!5} \multirow{-2}{*}{\cellcolor{white}\textbf{300W}}
& \textbf{+SB} & \textbf{100} & 2.81 / \textbf{2.79} & \textbf{0.1} & 71.2 \\
\midrule
\cellcolor{white} & Base & 98 & \textbf{3.97} & \textbf{2.0} & \textbf{60.8} \\
\rowcolor{green!5} \multirow{-2}{*}{\cellcolor{white}\textbf{WFLW}}
& \textbf{+SB} & \textbf{100} & 3.94 / \textbf{3.72} & 2.2 & 60.3 \\
\midrule
\cellcolor{white} & Base & \textbf{100} & \textbf{4.53} & \textbf{0.4} & \textbf{54.7} \\
\rowcolor{green!5} \multirow{-2}{*}{\cellcolor{white}\textbf{COFW}}
& \textbf{+SB} & \textbf{100} & 4.50 / \textbf{4.23} & \textbf{0.4} & 54.3 \\
\bottomrule
\end{tabular}
\end{minipage}%
\hfill
\begin{minipage}[t]{0.53\textwidth}
\centering
\footnotesize
\renewcommand{\arraystretch}{1.1}
\setlength{\tabcolsep}{2.5pt}
\caption{\textbf{SOTA comparison on original test sets.} Landmark localization error (NME$_{\text{all}}$\,$\downarrow$).}
\label{tab:sota_original}
\begin{tabular}{lccc}
\toprule
\cellcolor{blue!10}\textbf{Method} &
\cellcolor{cyan!10}\textbf{300W} &
\cellcolor{teal!12}\textbf{WFLW} &
\cellcolor{green!12}\textbf{COFW} \\
\midrule
AWing~\citep{wang2019adaptive} & 3.07 & 4.36 & 4.94 \\
LUVLi~\citep{kumar2020luvli} & 3.23 & 4.37 & -- \\
ViTPose~\citep{xu2022vitpose} & 3.02 & 4.26 & 4.85 \\
ADNet~\citep{zhu2022reasoning} & 2.93 & 4.14 & 4.68 \\
HIH~\citep{zhu2022occlusion} & 3.09 & 4.08 & 4.63 \\
SLPT~\citep{li2024cascaded} & 3.17 & 4.14 & 4.79 \\
STAR~\citep{zhou2023star} & 2.90 & 4.03 & 4.62 \\
OccFace~\citep{xiang2026occface} & 2.83 & 3.77 & 4.53 \\
\midrule
HR-noSA~\citep{yang2025heatmap} & 2.86 & 3.97 & 4.53 \\
\rowcolor{green!5} \quad \textbf{+ \sbayes} (mean) & 2.81 & 3.94 & 4.50 \\
\rowcolor{green!5} \quad \textbf{$\hookrightarrow$ minNME$_{@100}$} & \textbf{2.79} & \textbf{3.72} & \textbf{4.23} \\
\bottomrule
\end{tabular}
\end{minipage}
\end{table*}

\noindent \textbf{Original Test Sets and SOTA Comparison.} On clean images, predicted uncertainties vanish ($\sigma_i^2 \to 0$), causing observation precision ($W_i = \sigma_i^{-2}$) to dominate the latent prior. Consequently, \sbayes acts as an identity mapping ($\boldsymbol{\mu}_{\theta} \approx \hat{\mathbf{y}}$), preserving or marginally improving base accuracy (e.g., 300W NME $2.86 \rightarrow 2.81$ in Tables~\ref{tab:softargmax_original} and~\ref{tab:sota_original}). Furthermore, drawing 100 posterior samples yields SOTA accuracy ($\text{minNME}_{@100}$), outperforming specialized models like OccFace and ViTPose. This confirms that the posterior space contains valid and accurate geometric hypotheses.

\begin{keyresult}{Key Result 2: SOTA Shape Accuracy}
Trusting visual evidence where confident and shape priors where uncertain, \sbayes improves accuracy while preserving structural integrity.
\end{keyresult}

\begin{figure}[t]
\centering
\begin{minipage}[c]{0.45\textwidth}
\centering
\small
\renewcommand{\arraystretch}{1.2}
\setlength{\tabcolsep}{5pt}
\captionof{table}{\textbf{Uncertainty evaluation} on occluded 300W. $^*$Native uncertainty estimates from the base model.}
\label{tab:calib_300w_wrapped}
\begin{tabular}{l cc cc}
\toprule
\rowcolor{blue!12} & \multicolumn{2}{c}{\textbf{NLL\,$\downarrow$}} & \multicolumn{2}{c}{\textbf{ECE\,$\downarrow$}} \\
\rowcolor{blue!12} \textbf{Base Model} & Base & \textbf{+SB} & Base & \textbf{+SB} \\
\midrule
LUVLi$^*$ & 6.08 & \textbf{6.07} & 0.90 & \textbf{0.86} \\
HR-noSA   & 6.48 & \textbf{6.34} & 7.56 & \textbf{1.97} \\
ORFormer  & 6.50 & \textbf{6.35} & 4.87 & \textbf{2.07} \\
PiPNet    & 6.95 & \textbf{6.55} & 3.36 & \textbf{1.09} \\
STAR      & 6.81 & \textbf{6.60} & 7.10 & \textbf{1.95} \\
\bottomrule
\end{tabular}
\end{minipage}%
\hfill
\begin{minipage}[c]{0.5\textwidth}
\centering
\includegraphics[width=\linewidth]{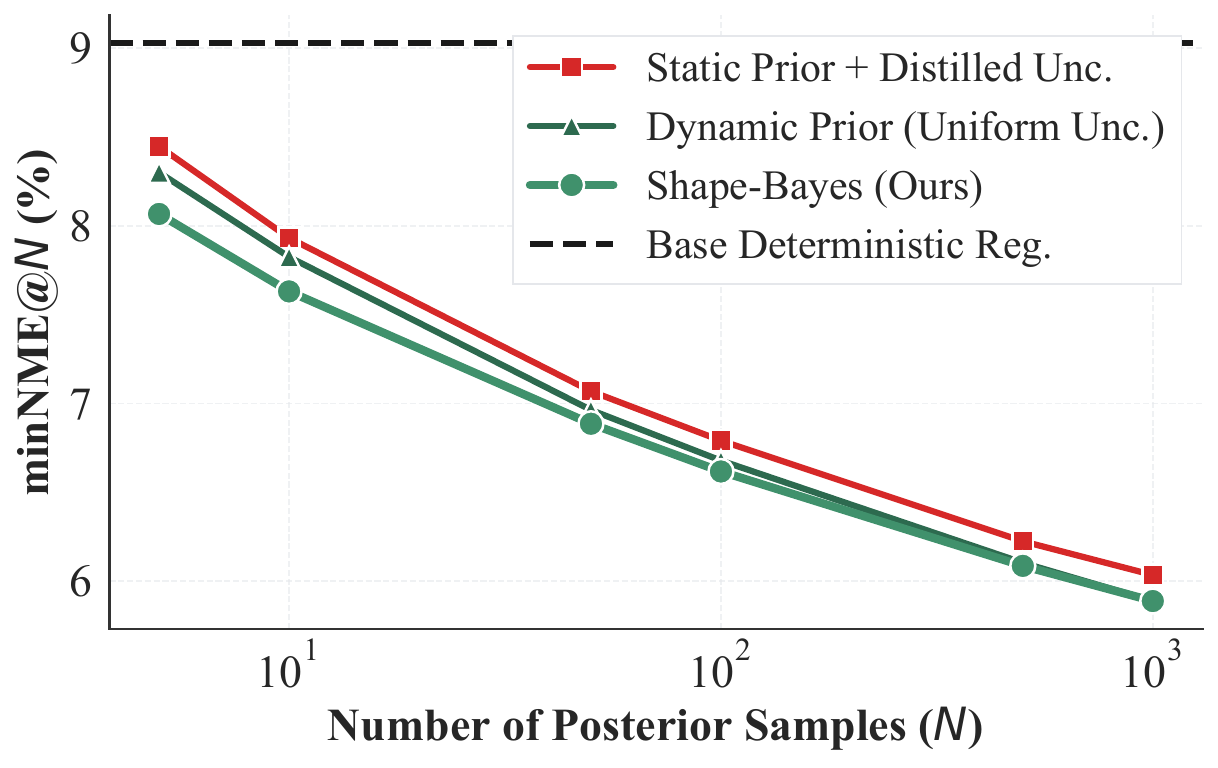}
\caption{\textbf{minNME$_{@N}$ ablation} on occluded 300W as a function of the number of posterior samples $N$.}
\label{fig:ablation_n_samples}
\end{minipage}
\end{figure}

\noindent \textbf{Posterior Uncertainty Calibration.} Well-calibrated uncertainty estimates are vital for downstream applications where detecting a prediction failure is as critical as shape accuracy. Table~\ref{tab:calib_300w_wrapped} evaluates calibration metrics across base architectures on the occluded 300W benchmark. Under out-of-distribution dynamic occlusions, most base models exhibit severe overconfidence, yielding high Expected Calibration Error (ECE) scores (e.g., 7.56 for HR-noSA and 7.10 for STAR). The notable exception is LUVLi, which achieves better baseline calibration via explicit supervision of landmark covariance. \sbayes effectively mitigates this widespread overconfidence: analytic posterior updates consistently reduce ECE across all architectures. For instance, HR-noSA's ECE drops substantially to 1.97, while LUVLi's already low ECE further decreases to 0.86. Consistent improvements in Negative Log-Likelihood (NLL) confirm that the \sbayes posteriors provide fundamentally more reliable confidence bounds for shape inference under visual ambiguity.

\begin{figure*}[t]
    \centering
    \includegraphics[width=\textwidth]{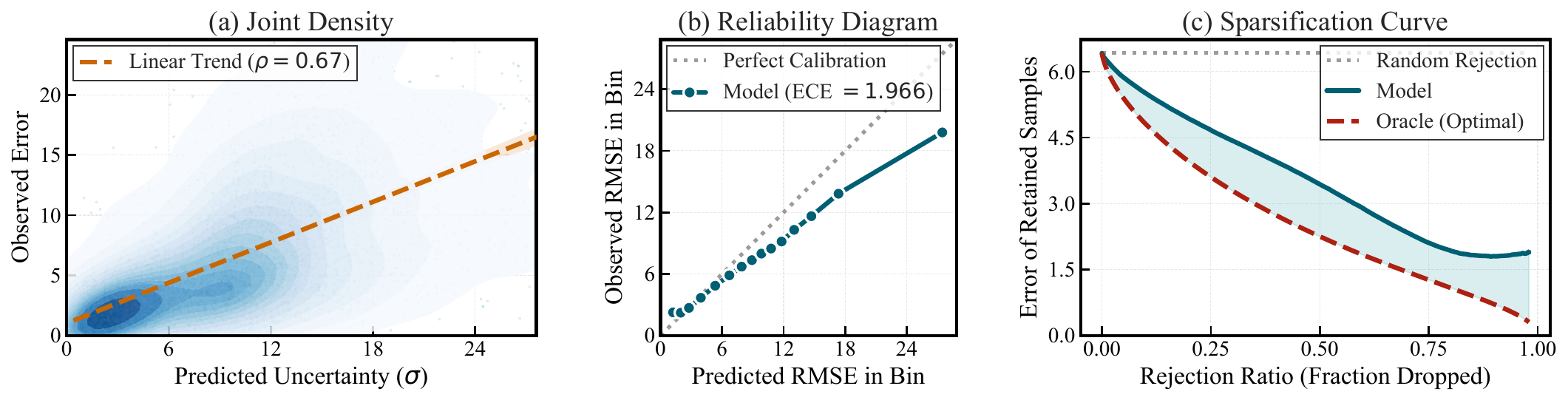}
    \caption{\textbf{Posterior calibration} on occluded 300W. (a) Error vs.\ uncertainty shows strong correlation ($\rho=0.67$). (b) Reliability plot yields robust calibration ($\text{ECE}=1.97$). (c) Sparsification curve tracks the oracle.}
    \label{fig:uncertainty_calibration_300w}
    \vspace{-4mm}
\end{figure*}

\sbayes achieves this improved posterior uncertainty calibration by mapping unconstrained predictions onto a structured latent manifold, tightly aligning the predicted variance with actual empirical error. Consequently, high aleatoric uncertainty naturally propagates into a wider Gaussian posterior variance, inherently preventing overly confident mispredictions. As Figure~\ref{fig:uncertainty_calibration_300w} shows, the estimated aleatoric uncertainty correlates strongly with observed structural errors (Spearman $\rho = 0.67$). Moreover, the sparsification curve demonstrates that progressively filtering these uncertain predictions successfully isolates highly accurate shapes, closely tracking the oracle lower bound.

\begin{keyresult}{Key Result 3: Well-Calibrated Shape Posterior Distributions}
By propagating aleatoric uncertainty through a prior-constrained Bayesian update, \sbayes infers well-calibrated posteriors.
\end{keyresult}

\noindent \textbf{Ablation Study.} Table~\ref{tab:ablations_occlusion} details the architectural progression of \sbayes on occluded 300W (all metrics computed on occluded landmarks). Starting from the unconstrained \textit{Base Deterministic Regression} baseline using HR-noSA ($\text{NME} = 9.03$), applying a naive \textit{Deterministic Manifold Projection} with PCA using unweighted least-squares yields a modest error reduction ($\text{NME} = 8.85$). Because standard least-squares assumes isotropic, homogeneous noise ($\mathbf{\Sigma} = \sigma^2 \mathbf{I}$), corrupted landmarks act as heavy outliers that distort the shape manifold and pull confident landmarks out of place.

Integrating variance weights resolves this leverage issue (\textit{Static Prior + Distilled Aleatoric Uncertainty}). Weighting landmarks by inverse variance transforms unweighted Euclidean projection into Mahalanobis distance minimization. Severely occluded points are automatically down-weighted, allowing the structural prior to complete their positions. This variance weighting lowers NME from $8.85$ to $8.08$ and reduces FR from $31.57\%$ to $25.69\%$. Crucially, posterior distributions enable hypothesis sampling, yielding a substantially lower $\text{minNME}_{@100}$ of $6.79$ (Figure~\ref{fig:ablation_n_samples}).

Finally, replacing the static prior with an adaptive dynamic prior yields the complete \sbayes \textit{(Dynamic Prior + Distilled Unc.)} framework. While a static prior can be overly rigid, the dynamic prior adaptively accommodates input-specific variations (e.g., head pose of face shapes). As Table~\ref{tab:ablations_occlusion} shows, this unified approach achieves the best performance across all evaluation criteria: an NME of $7.91$, $\text{minNME}_{@100}$ of $6.60$, FR of $24.67\%$, and AUC of $31.53\%$.

\begin{table}[!htb]
\centering
\small
\renewcommand{\arraystretch}{1.1}
\setlength{\tabcolsep}{5pt}
\caption{\textbf{Ablation of \sbayes on occluded 300W.} Rows are ordered by architectural progression. Cell shading tracks performance from high failure rate (\textcolor{red}{red}) to low failure rate (\textcolor{blue}{blue}).}
\label{tab:ablations_occlusion}
\begin{tabular}{lcccc}
\toprule
\textbf{Architectural Configuration} & \textbf{NME\,$\downarrow$} & \textbf{minNME$_{@100}$\,$\downarrow$} & \textbf{FR\,$\downarrow$} & \textbf{AUC\,$\uparrow$} \\
\midrule
Base Deterministic Regression & \cellcolor{red!15}9.03 & \cellcolor{red!15}9.03 & \cellcolor{red!15}32.15\% & \cellcolor{red!15}25.81\% \\
Deterministic Manifold Projection & \cellcolor{red!5}8.85 & \cellcolor{red!5}8.85 & \cellcolor{red!5}31.57\% & \cellcolor{red!5}26.97\% \\
\midrule
Static Prior + Distilled Aleatoric Uncertainty & \cellcolor{blue!5}8.08 & \cellcolor{blue!5}6.79 & \cellcolor{blue!5}25.69\% & \cellcolor{blue!5}30.61\% \\
Dynamic Prior (w/ Uniform Uncertainty) & \cellcolor{blue!10}8.00 & \cellcolor{blue!10}6.68 & \cellcolor{blue!10}25.47\% & \cellcolor{blue!10}30.80\% \\
\textbf{\sbayes (Dynamic Prior + Distilled Unc.)} & \cellcolor{blue!20}\textbf{7.91} & \cellcolor{blue!20}\textbf{6.60} & \cellcolor{blue!20}\textbf{24.67\%} & \cellcolor{blue!20}\textbf{31.53\%} \\
\bottomrule
\end{tabular}
\end{table}

\FloatBarrier

%% file: sections/related_work.tex
\section{Related Work}
\label{sec:related_work}
\noindent \textbf{Probabilistic Shape Inference.} Classical computer vision relied on linear Gaussian subspaces, such as Active Shape Models~\citep{cootes1995active}, to enforce geometric consistency. While modern generative models~\citep{kingma2013auto, xu2024bayesian} capture complex priors, they require expensive sampling and struggle to analytically ingest localized uncertainty. Concurrently, PPL~\citep{wang2026pose} constructs categorical priors via discrete memory banks of prototypical poses, requiring iterative refinement. In contrast, \sbayes couples a Transformer-encoded prior with shape uncertainty, yielding an exact, single-pass Gaussian update for structured shapes.

\noindent \textbf{Uncertainty Modeling in Shape Inference.}
While existing shape regressors parameterize coordinate uncertainty via residual likelihoods~\citep{li2021human}, Dirichlet mapping~\citep{neven2021dirichlet}, adaptive heatmaps~\citep{kumar2020luvli}, or conformal sets~\citep{jonkers2025reliable} for passive tasks like filtering, \sbayes actively exploits it. Paralleling recent work on transformation-equivariant variance~\citep{tian2025test}, we distill affine-augmented teacher uncertainty into base regressors. Crucially, we inject this uncertainty into a closed-form Bayesian update, elevating uncertainty from a passive metric to the core mechanism arbitrating between noisy local predictions and global priors.

\noindent \textbf{Shape Inference under Occlusion.}
Standard shape regression exhibits structural collapse under severe occlusion. While recent methods attempt explicit occlusion modeling~\citep{kumar2020luvli}, autoregressive joint prediction~\citep{zheng2025hipart}, or spatial prior propagation~\citep{zheng2026pandapose}, alongside relational reasoning~\citep{zhu2022reasoning, xiang2026occface, chiang2025orformer}, they lack tightly coupled global shape constraints. Consequently, once local feature evidence degrades, unconstrained regression or autoregressive formulations lack a mechanism to enforce global geometric bounds, causing their predictions to drift outside the valid shape manifold. In contrast, \sbayes performs analytical Gaussian conditioning, rigorously fusing heteroscedastic observations with a conditional shape prior in closed form to guarantee structural validity by construction.

%% file: sections/conclusion.tex
\section{Conclusion}
\label{sec:conclusion}

We presented \sbayes, a probabilistic framework for shape inference under visual ambiguity. By unifying distilled aleatoric uncertainty with a learned conditional prior, it computes exact closed-form posteriors. This explicitly balances noisy local evidence against global structural constraints, preventing structural collapse under severe occlusion. Benchmarked across six face alignment base architectures, \sbayes consistently reduces shape error, sharpens uncertainty calibration, and enforces structural validity, providing a principled solution for structured shape inference.

%% file: sections/appendix.tex
\section{Qualitative Results on Occluded Image Test Sets}
\label{app:qualitative}

\begin{figure}[!htbp]
    \centering
    \includegraphics[width=\textwidth,height=0.82\textheight,keepaspectratio]{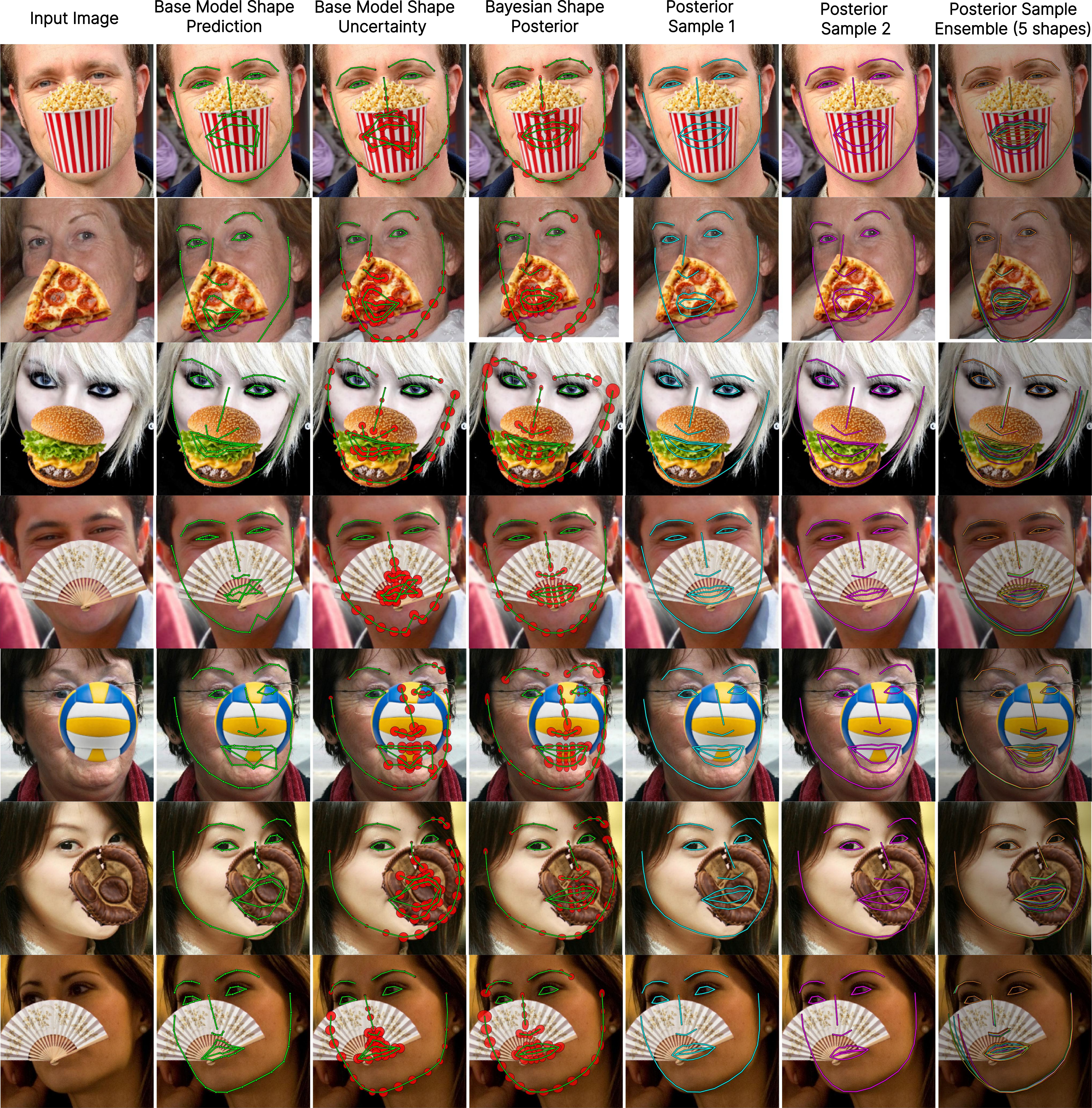}
    \caption{\textbf{Qualitative results on occluded 300W.} HR-noSA~\citep{yang2025heatmap} base model vs.\ \sbayes.}
    \label{fig:300w_qual}
\end{figure}

\begin{figure}[!htbp]
    \centering
    \includegraphics[width=\textwidth,height=0.82\textheight,keepaspectratio]{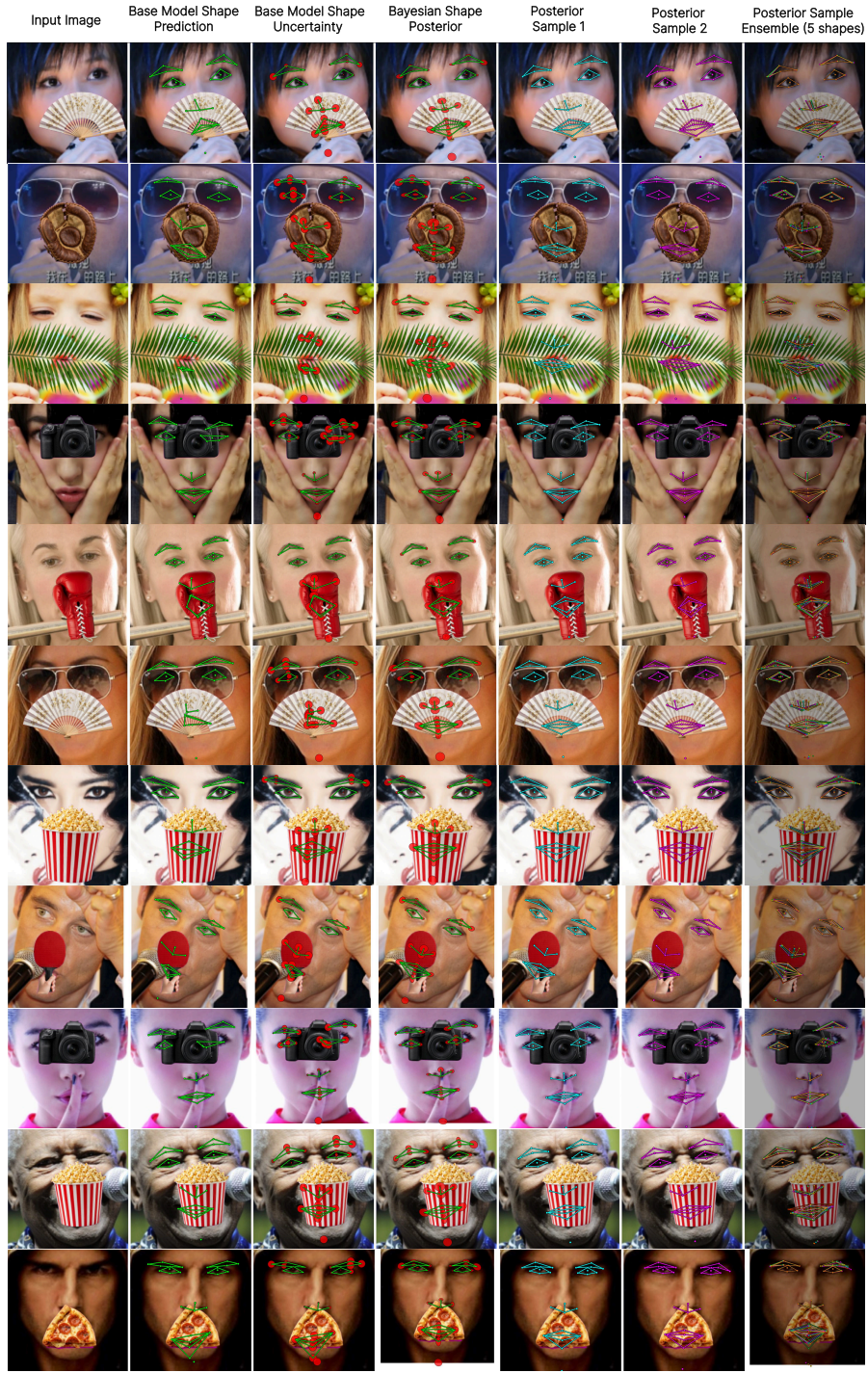}
    \caption{\textbf{Qualitative results on occluded COFW.} HR-noSA~\citep{yang2025heatmap} base model vs.\ \sbayes.}
    \label{fig:cofw_qual}
\end{figure}

\begin{figure}[!htbp]
    \centering
    \includegraphics[width=\textwidth,height=0.82\textheight,keepaspectratio]{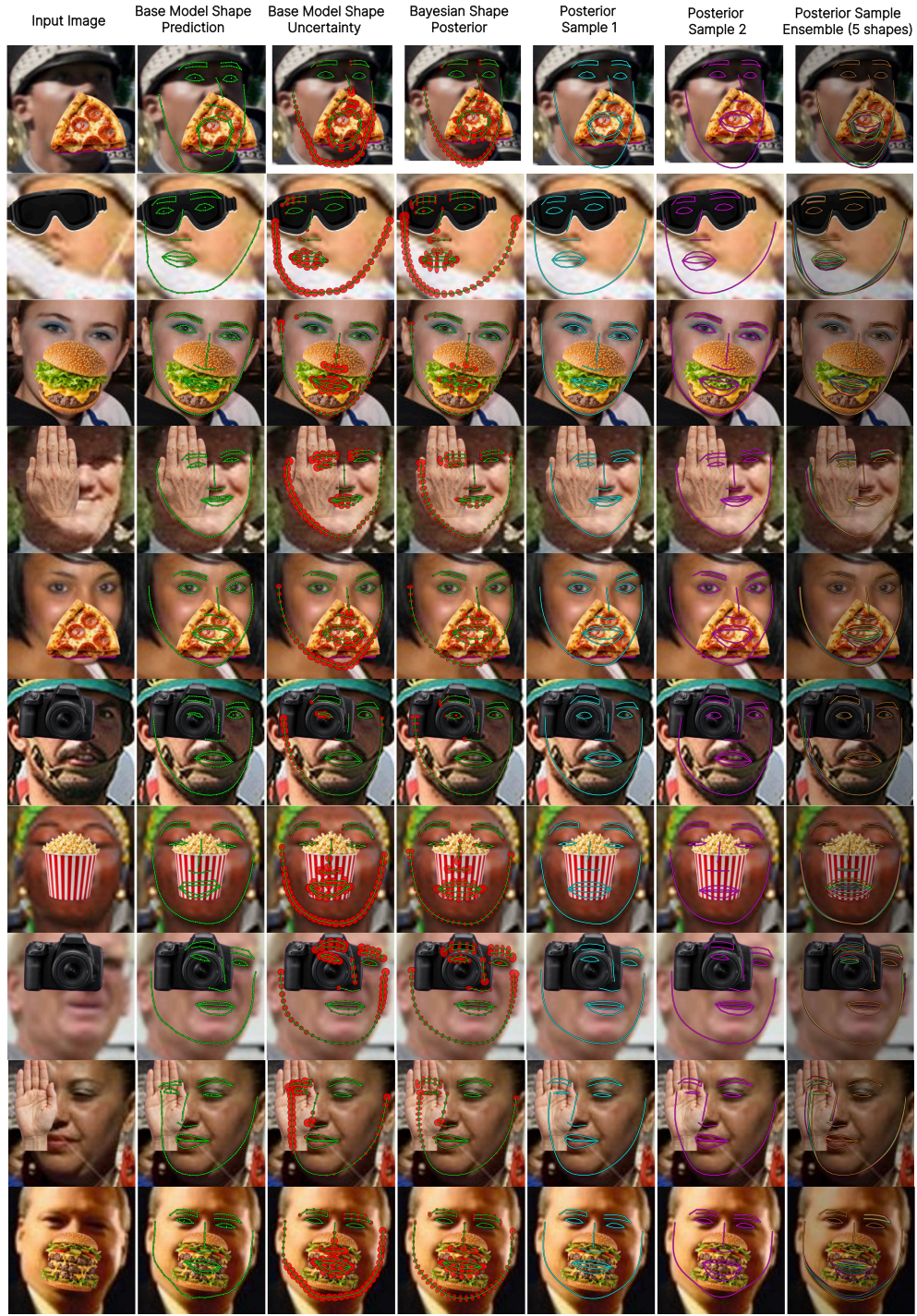}
    \caption{\textbf{Qualitative results on occluded WFLW.} HR-noSA~\citep{yang2025heatmap} base model vs.\ \sbayes.}
    \label{fig:wflw_qual}
\end{figure}

Figures~\ref{fig:300w_qual},~\ref{fig:cofw_qual}, and~\ref{fig:wflw_qual} present qualitative comparisons between HR-noSA~\citep{yang2025heatmap}, a current state-of-the-art deterministic regression model, and its \sbayes extension on dynamically occluded test images from all three datasets. In each case, the base model exhibits characteristic structural collapse: landmarks within occluded regions drift to implausible locations, producing anatomically incoherent face shapes. \sbayes corrects these failures by projecting the noisy observations onto the learned shape manifold via the closed-form Bayesian update, recovering plausible facial geometries even under severe occlusion.

\FloatBarrier

\section{Region-Wise Shape Accuracy Improvements Analysis}
\label{app:regional_nme}

To verify that the \sbayes manifold constraint does not introduce regional bias, for instance, improving jaw predictions at the expense of eye localization, we decompose the occluded NME into anatomically distinct facial regions. Tables~\ref{tab:reg_300w},~\ref{tab:reg_wflw}, and~\ref{tab:reg_cofw} present this breakdown for each benchmark. Across all three datasets and all base models, \sbayes delivers consistent NME reductions in every facial region, confirming that the Bayesian manifold formulation regularizes coordinate estimates globally rather than trading off accuracy between components.

\begin{table}[H]
\centering
\small
\renewcommand{\arraystretch}{1.2}
\setlength{\tabcolsep}{4pt}
\caption{{Region-wise shape prediction error in terms of NME$_{\text{occ}}$}($\downarrow$) on \textbf{occluded 300W}.}
\label{tab:reg_300w}
\begin{tabular}{l l c c c c c c c}
\toprule
\rowcolor{HeaderBg} \textbf{Base Model} & \textbf{Config.} & \textbf{Jaw} & \textbf{R-Brow} & \textbf{L-Brow} & \textbf{Nose} & \textbf{L-Eye} & \textbf{R-Eye} & \textbf{Mouth} \\
\midrule
\multirow{2}{*}{\textbf{ORFormer~\citep{chiang2025orformer}}}
  & Base & 9.0 & 9.5 & 9.8 & 9.5 & 7.1 & 6.6 & 11.1 \\
  & \textbf{+SB} & \best{8.8} & \best{8.8} & \best{9.3} & \best{7.8} & \best{6.3} & \best{5.7} & \best{9.7} \\
\midrule
\multirow{2}{*}{\textbf{LUVLi~\citep{kumar2020luvli}}}
  & Base & 10.7 & 14.1 & 12.9 & 9.0 & 10.2 & 9.3 & 9.3 \\
  & \textbf{+SB} & \best{9.9} & \best{11.3} & \best{11.1} & \best{7.4} & \best{8.0} & \best{7.7} & \best{8.8} \\
\midrule
\multirow{2}{*}{\textbf{STAR~\citep{zhou2023star}}}
  & Base & 13.0 & 10.2 & 12.9 & 10.5 & 9.1 & 7.2 & 12.0 \\
  & \textbf{+SB} & \best{11.5} & \best{9.2} & \best{10.6} & \best{7.7} & \best{7.3} & \best{6.1} & \best{10.2} \\
\midrule
\multirow{2}{*}{\textbf{PIPNet~\citep{jin2021pixel}}}
  & Base & 16.7 & 16.9 & 18.0 & 11.7 & 11.2 & 11.1 & 15.6 \\
  & \textbf{+SB} & \best{13.6} & \best{12.0} & \best{12.7} & \best{9.3} & \best{7.7} & \best{7.6} & \best{12.6} \\
\bottomrule
\end{tabular}
\end{table}

\begin{table}[H]
\centering
\small
\renewcommand{\arraystretch}{1.2}
\setlength{\tabcolsep}{4pt}
\caption{{Region-wise shape prediction error in terms of NME$_{\text{occ}}$}($\downarrow$) on \textbf{occluded WFLW}.}
\label{tab:reg_wflw}
\begin{tabular}{l l c c c c c c c}
\toprule
\rowcolor{HeaderBg} \textbf{Base Model} & \textbf{Config.} & \textbf{Jaw} & \textbf{R-Brow} & \textbf{L-Brow} & \textbf{Nose} & \textbf{L-Eye} & \textbf{R-Eye} & \textbf{Mouth} \\
\midrule
\multirow{2}{*}{\textbf{HR-noSA~\citep{yang2025heatmap}}}
  & Base & 11.8 & 11.9 & 11.4 & 11.3 & 9.6 & 10.4 & 12.9 \\
  & \textbf{+SB} & \best{11.5} & \best{11.0} & \best{10.9} & \best{9.9} & \best{8.9} & \best{9.3} & \best{11.8} \\
\midrule
\multirow{2}{*}{\textbf{ORFormer~\citep{chiang2025orformer}}}
  & Base & 11.5 & 11.2 & 11.4 & 9.0 & 9.5 & 8.7 & 10.5 \\
  & \textbf{+SB} & \best{11.0} & \best{10.9} & \best{11.2} & \best{8.4} & \best{9.1} & \best{8.4} & \best{10.2} \\
\midrule
\multirow{2}{*}{\textbf{LUVLi~\citep{kumar2020luvli}}}
  & Base & 11.5 & 12.9 & 13.6 & 11.3 & 11.5 & 11.1 & 12.8 \\
  & \textbf{+SB} & \best{11.1} & \best{12.2} & \best{12.7} & \best{10.2} & \best{10.5} & \best{10.2} & \best{11.8} \\
\midrule
\multirow{2}{*}{\textbf{STAR~\citep{zhou2023star}}}
  & Base & 11.4 & 10.4 & 10.2 & 10.6 & 8.9 & 8.6 & 12.6 \\
  & \textbf{+SB} & \best{11.2} & \best{10.0} & \best{9.8} & \best{9.6} & \best{8.3} & \best{8.2} & \best{11.8} \\
\midrule
\multirow{2}{*}{\textbf{DSLPT~\citep{xia2023robust}}}
  & Base & \best{11.3} & 11.6 & 12.1 & 11.6 & 9.7 & 9.1 & 12.9 \\
  & \textbf{+SB} & 11.4 & \best{11.2} & \best{11.6} & \best{10.1} & \best{9.3} & \best{8.8} & \best{12.2} \\
\midrule
\multirow{2}{*}{\textbf{PIPNet~\citep{jin2021pixel}}}
  & Base & 12.2 & 14.0 & 13.3 & 9.0 & 10.6 & 11.0 & 11.3 \\
  & \textbf{+SB} & \best{11.2} & \best{11.4} & \best{11.3} & \best{8.1} & \best{9.3} & \best{9.4} & \best{10.3} \\
\bottomrule
\end{tabular}
\end{table}

\begin{table}[H]
\centering
\small
\renewcommand{\arraystretch}{1.2}
\setlength{\tabcolsep}{4pt}
\caption{{Region-wise shape prediction error in terms of NME$_{\text{occ}}$}($\downarrow$) on \textbf{occluded COFW}.}
\label{tab:reg_cofw}
\begin{tabular}{l l c c c c c c c}
\toprule
\rowcolor{HeaderBg} \textbf{Base Model} & \textbf{Config.} & \textbf{R-Brow} & \textbf{L-Brow} & \textbf{R-Eye} & \textbf{L-Eye} & \textbf{Nose} & \textbf{Mouth} & \textbf{Chin} \\
\midrule
\multirow{2}{*}{\textbf{HR-noSA~\citep{yang2025heatmap}}}
  & Base & 12.0 & 14.4 & 10.0 & 12.1 & 10.7 & 11.1 & \best{13.4} \\
  & \textbf{+SB} & \best{10.6} & \best{11.7} & \best{8.5} & \best{8.9} & \best{9.1} & \best{9.8} & 14.6 \\
\midrule
\multirow{2}{*}{\textbf{ORFormer~\citep{chiang2025orformer}}}
  & Base & 10.1 & 11.6 & 8.0 & 8.8 & 9.3 & 9.9 & 8.7 \\
  & \textbf{+SB} & \best{9.0} & \best{10.4} & \best{7.5} & \best{7.9} & \best{8.6} & \best{9.5} & \best{8.1} \\
\midrule
\multirow{2}{*}{\textbf{STAR~\citep{zhou2023star}}}
  & Base & 11.9 & 14.4 & 9.9 & 12.0 & 10.3 & 10.1 & 13.0 \\
  & \textbf{+SB} & \best{10.7} & \best{12.5} & \best{8.6} & \best{9.9} & \best{9.0} & \best{9.3} & \best{12.8} \\
\midrule
\multirow{2}{*}{\textbf{PIPNet~\citep{jin2021pixel}}}
  & Base & 12.1 & 13.3 & 11.2 & 10.5 & 10.0 & 10.2 & 12.0 \\
  & \textbf{+SB} & \best{10.9} & \best{11.4} & \best{9.0} & \best{8.4} & \best{9.0} & \best{9.8} & \best{11.6} \\
\bottomrule
\end{tabular}
\end{table}

\FloatBarrier

\section{Shape-Bayes Transformer: Computational Complexity Analysis}
\label{app:complexity}

We provide a detailed breakdown of the computational cost of the \sbayes Transformer encoder and the closed-form Bayesian solve, benchmarked against the StackedHGNetV1 network used in HR-noSA~\citep{yang2025heatmap}. All figures are derived from the 300W configuration ($N = 68$ landmarks, $K = 60$ PCA components) and verified against PyTorch parameter counts.

\subsection{Shape-Bayes Transformer Architecture}

The transformer encoder maps per-landmark observation pairs $(\boldsymbol{s}_i, \boldsymbol{\sigma}_i) \in \mathbb{R}^{4}$ for $i = 1, \ldots, N$ to a $K$-dimensional log-precision vector. Its hyperparameters are: embedding dimension $d = 64$, feedforward dimension $d_{\text{ff}} = 128$, $L = 5$ encoder layers, and $H = 8$ attention heads (head dimension $d_h = d / H = 8$).

\subsection{Parameter Count}

\begin{table}[H]
\centering
\caption{Shape-Bayes Transformer parameter breakdown (300W, $N\!=\!68$, $K\!=\!60$, $d\!=\!64$, $d_{\text{ff}}\!=\!128$, $L\!=\!5$, $H\!=\!8$). All counts verified against \texttt{torch.nn.TransformerEncoder}.}
\label{tab:params}
\small
\begin{tabular}{lrrr}
\toprule
\textbf{Component} & \textbf{Formula} & \textbf{Per-Unit} & \textbf{Subtotal} \\
\midrule
\multicolumn{4}{l}{\textit{Input Projection}} \\
\quad \texttt{Linear}$(4 \to d)$ & $4d + d$ & 320 & 320 \\
\midrule
\multicolumn{4}{l}{\textit{Per Transformer Encoder Layer ($\times L = 5$)}} \\
\quad Multi-head self-attention & & & \\
\quad\quad In-projection $W_{QKV}$ & $3d^2 + 3d$ & 12{,}480 & \\
\quad\quad Out-projection $W_O$ & $d^2 + d$ & 4{,}160 & \\
\quad Self-attention subtotal & & 16{,}640 & \\[2pt]
\quad Position-wise feedforward & & & \\
\quad\quad \texttt{Linear}$(d \to d_{\text{ff}})$ & $d \cdot d_{\text{ff}} + d_{\text{ff}}$ & 8{,}320 & \\
\quad\quad \texttt{Linear}$(d_{\text{ff}} \to d)$ & $d_{\text{ff}} \cdot d + d$ & 8{,}256 & \\
\quad Feedforward subtotal & & 16{,}576 & \\[2pt]
\quad LayerNorm $\times 2$ & $2 \times 2d$ & 256 & \\
\cmidrule{2-4}
\quad \textbf{Per-layer total} & & \textbf{33{,}472} & \textbf{167{,}360} \\
\midrule
\multicolumn{4}{l}{\textit{Post-Encoder}} \\
\quad Final LayerNorm & $2d$ & 128 & 128 \\
\quad Precision head \texttt{Linear}$(Nd \to K)$ & $N \cdot d \cdot K + K$ & 261{,}180 & 261{,}180 \\
\midrule
\textbf{Total} & & & \textbf{428{,}988} \\
\bottomrule
\end{tabular}
\end{table}

\paragraph{Remark.} The precision head dominates the parameter budget (60.9\% of total), a direct consequence of flattening all $N$ token embeddings before the final linear projection. The five transformer layers collectively contribute only 167{,}360 parameters (39.0\%).

\subsection{FLOPs Analysis}
\label{app:flops}

We count floating-point operations (FLOPs) using the standard convention where one multiply-accumulate equals two FLOPs, i.e., a matrix multiplication $(M, P) \times (P, Q)$ contributes $2MPQ$ FLOPs.

\begin{table}[H]
\centering
\caption{Per-layer FLOPs breakdown for the \sbayes Transformer ($N\!=\!68$, $d\!=\!64$, $d_{\text{ff}}\!=\!128$, $H\!=\!8$).}
\label{tab:flops}
\small
\begin{tabular}{lrr}
\toprule
\textbf{Operation} & \textbf{Formula} & \textbf{FLOPs} \\
\midrule
\multicolumn{3}{l}{\textit{Multi-Head Self-Attention}} \\
\quad $Q, K, V$ projections & $2 \cdot N \cdot d \cdot 3d$ & 1{,}671{,}168 \\
\quad Attention scores $QK^\top$ & $2 \cdot N^2 \cdot d$ & 591{,}872 \\
\quad Attention $\times$ values & $2 \cdot N^2 \cdot d$ & 591{,}872 \\
\quad Output projection & $2 \cdot N \cdot d^2$ & 557{,}056 \\
\cmidrule{2-3}
\quad Self-attention subtotal & $2N(4d^2 + 2Nd)$ & 3{,}411{,}968 \\
\midrule
\multicolumn{3}{l}{\textit{Position-Wise Feedforward}} \\
\quad \texttt{Linear}$(d \to d_{\text{ff}})$ + \texttt{Linear}$(d_{\text{ff}} \to d)$ & $4 \cdot N \cdot d \cdot d_{\text{ff}}$ & 2{,}228{,}224 \\
\midrule
\textbf{Per-layer total} & & \textbf{5{,}640{,}192} \\
\bottomrule
\end{tabular}
\end{table}

\noindent Summing across all stages:

\begin{table}[H]
\centering
\caption{End-to-end FLOPs for the complete \sbayes inference pipeline.}
\label{tab:total_flops}
\small
\begin{tabular}{lrr}
\toprule
\textbf{Stage} & \textbf{Formula} & \textbf{FLOPs} \\
\midrule
Input projection & $2 \cdot N \cdot 4 \cdot d$ & 34{,}816 \\
Transformer encoder ($L\!=\!5$ layers) & $L \times 5.64\text{M}$ & 28{,}200{,}960 \\
Precision head & $2 \cdot (Nd) \cdot K$ & 522{,}240 \\
\cmidrule{2-3}
\textit{Neural network subtotal} & & 28{,}758{,}016 \\
\midrule
\multicolumn{3}{l}{\textit{Closed-Form Bayesian Solve (Section~\ref{sec:method})}} \\
\quad $\mathbf{P}^\top \operatorname{diag}(\mathbf{w})$ scaling & $KM$ & 8{,}160 \\
\quad $(\mathbf{P}^\top \mathbf{W}) \mathbf{P}$ matrix product & $2K^2 M$ & 979{,}200 \\
\quad $(\mathbf{P}^\top \mathbf{W}) (\mathbf{y} - \bar{\mathbf{s}})$ & $2KM$ & 16{,}320 \\
\quad Cholesky factorisation + triangular solves$^{\dagger}$ & $\tfrac{1}{3}K^3 + 2K^2$ & 79{,}200 \\
\cmidrule{2-3}
\textit{Bayesian solve subtotal (MAP)} & & 1{,}083{,}484 \\
\quad Full matrix inverse (for posterior sampling)$^{\dagger\dagger}$ & $\tfrac{7}{3}K^3$ & $+$432{,}000 \\
\textit{Bayesian solve subtotal (MAP + sampling covariance)} & & 1{,}515{,}484 \\
\midrule
\textbf{Grand total (MAP path)} & & $\mathbf{\sim\!29.8\textbf{M}}$ \\
\textbf{Grand total (with posterior sampling)} & & $\mathbf{\sim\!30.3\textbf{M}}$ \\
\bottomrule
\end{tabular}
\end{table}

\noindent where $M = 2N = 136$ is the flattened landmark dimension. \\[4pt]
{\footnotesize $^{\dagger}$\,Computing the MAP estimate $\mathbf{b}_\mu = \tilde{\mathbf{\Sigma}}_b \mathbf{P}^T \mathbf{W} (\mathbf{s}' - \bar{\mathbf{S}})$ via \texttt{torch.linalg.solve} requires only Cholesky factorisation ($\tfrac{1}{3}K^3$) and two triangular solves ($2K^2$).} \\
{\footnotesize $^{\dagger\dagger}$\,Posterior sampling from $\mathcal{N}(\hat{\mathbf{b}}, \boldsymbol{\Lambda}^{-1})$ additionally requires the full matrix inverse, adding $\tfrac{7}{3}K^3$ FLOPs to recover $\boldsymbol{\Lambda}^{-1}$ via back-substitution against the identity. The dominant cost in both paths remains the $(\mathbf{P}^\top \mathbf{W})\mathbf{P}$ matrix product (979K FLOPs, $\sim$90\%).}

\subsection{Comparison with the Base Landmark Regression Network}
\label{app:baseline_comparison}

The base model is a StackedHGNetV1~\citep{yang2025heatmap}: a 4-stack, 4-level stacked hourglass network with CoordConv and Adaptive Attention Masks (AAM), operating on $256 \times 256$ input images. Feature maps are processed at $64 \times 64$ spatial resolution through recursive hourglass modules with bottleneck residual blocks ($C = 256$, mid-channel $= 128$).

\begin{table}[H]
\centering
\caption{Computational cost comparison: \sbayes module vs.\ StackedHGNetV1 base model.}
\label{tab:comparison}
\small
\begin{tabular}{lrrr}
\toprule
\textbf{Model} & \textbf{Parameters} & \textbf{FLOPs} & \textbf{Relative Cost} \\
\midrule
StackedHGNetV1 (base) & $\sim$13.4M & $\sim$29.1\,GFLOPs & 1.00$\times$ \\
\sbayes module (MAP) & 429K & $\sim$29.8\,MFLOPs & -- \\
\sbayes module (with sampling) & 429K & $\sim$30.3\,MFLOPs & -- \\
\midrule
\textbf{Shape-Bayes / Base} & \textbf{3.2\%} & \textbf{0.10\%} & $\sim$$1000\times$ cheaper \\
\bottomrule
\end{tabular}
\end{table}

\paragraph{Key observations.}
\begin{enumerate}
    \item \textbf{Parameter overhead is marginal.} The \sbayes Transformer adds 429K parameters (3.2\% of the base network), all of which are in a separate module. The base model weights remain entirely frozen during \sbayes training and inference.
    \item \textbf{Compute overhead is negligible.} At $\sim$29.8--30.3M FLOPs (MAP to full posterior), the \sbayes module constitutes approximately 0.10\% of the base model's $\sim$29.1\,GFLOPs---a ${\sim}1000\times$ reduction. This cost is dominated by the five-layer transformer encoder (28.2M FLOPs, 94.6\%); the closed-form Bayesian solve contributes only 1.1--1.5M FLOPs ($\sim$3.6--5.1\%), of which the $\mathbf{P}^\top\mathbf{W}\mathbf{P}$ matrix product accounts for 979K FLOPs ($\sim$90\% of the solve).
    \item \textbf{The bottleneck is the base model, not \sbayes.} Self-attention over $N = 68$ tokens with $d = 64$ operates in the low-sequence, low-dimension regime ($N^2 d = 296{,}448$ per layer), where the quadratic attention cost is inconsequential. By contrast, the base model's hourglass modules process $64 \times 64 = 4{,}096$-token feature maps through 13 residual blocks per stack with $256$-channel convolutions.
    \item \textbf{Asymptotic complexity.} The transformer scales as $\mathcal{O}(L(N^2 d + N d \cdot d_{\text{ff}}))$ in FLOPs and $\mathcal{O}(L(d^2 + d \cdot d_{\text{ff}}) + NdK)$ in parameters. For typical face alignment settings ($N \leq 98$, $K \leq 80$), both remain well below 1M FLOPs and 1M parameters, respectively, ensuring the module generalises to denser landmark configurations (e.g., WFLW with $N = 98$) without meaningful cost increases.
\end{enumerate}

\paragraph{Wall-clock latency.} On a single NVIDIA RTX 3090, the \sbayes forward pass (transformer + Bayesian solve) adds $< 0.5$\,ms to the per-image inference time, which is dominated by the base model's convolutional backbone ($\sim$12\,ms). The posterior sampling step (\texttt{np.random.multivariate\_normal}) is similarly negligible at $< 0.1$\,ms per sample.

\subsection{Memory Footprint}

\begin{table}[H]
\centering
\caption{Inference-time memory overhead of the \sbayes module (FP32, batch size 1).}
\label{tab:memory}
\small
\begin{tabular}{lrl}
\toprule
\textbf{Component} & \textbf{Size} & \textbf{Notes} \\
\midrule
Model weights & 1.64\,MB & $428{,}988 \times 4$ bytes \\
PCA basis $\mathbf{P} \in \mathbb{R}^{M \times K}$ & 32.6\,KB & $136 \times 60 \times 4$ bytes, stored once \\
Mean shape $\bar{\mathbf{s}} \in \mathbb{R}^{M}$ & 0.53\,KB & $136 \times 4$ bytes \\
Activations (peak) & $\sim$170\,KB & Dominated by $L \times N \times d$ attention maps \\
Precision matrix $\boldsymbol{\Lambda} \in \mathbb{R}^{K \times K}$ & 14.4\,KB & Per-sample; used in \texttt{linalg.solve} \\
\midrule
\textbf{Total} & $<$ \textbf{2\,MB} & vs.\ $\sim$51\,MB for StackedHGNetV1 weights \\
\bottomrule
\end{tabular}
\end{table}

\section{Shape-Bayes Training and Optimization}
\label{app:shape_bayes_training}

The \sbayes Transformer encoder is trained offline using the stochastically augmented datasets (described in Appendix~\ref{app:training_augmentation}). The model processes the concatenated inputs of coordinate predictions $\mathbf{S}'$ and their associated uncertainties $\boldsymbol{\sigma}'$, yielding the diagonal prior precision matrix.

\noindent \textbf{Architecture Settings.} The network consists of a 5-layer Transformer encoder with an embedding dimension $d_{\text{model}} = 64$. The feed-forward network within each Transformer block has a hidden dimension of 128. Multi-head self-attention uses 8 heads, and a dropout rate of 0.1 is applied throughout to mitigate overfitting.

\noindent \textbf{Optimization and Hyperparameters.} The model is optimized end-to-end using the AdamW optimizer. We set the initial learning rate to $1 \times 10^{-3}$ and the weight decay to $1 \times 10^{-4}$. The learning rate schedule follows a Cosine Annealing with Warm Restarts strategy, configured with $T_0 = 1$ and $T_{\text{mult}} = 2$. The model is trained for 10,000 epochs with a batch size of 64 on the 300W and COFW datasets, and for 1,000 epochs with a batch size of 32 on the WFLW dataset.

The joint loss $\mathcal{L} = \lambda_{\text{MSE}} \mathcal{L}_{\text{MSE}} + \lambda_{\text{NLL}} \mathcal{L}_{\text{NLL}}$ uses $\lambda_{\text{MSE}} = 1000$ and $\lambda_{\text{NLL}} = 5 \times 10^{-4}$. The large MSE scaling compensates for gradient attenuation through the matrix inverse in the closed-form Bayesian update, while the NLL term operates with a stop-gradient on the posterior mean to ensure decoupled optimization of shape accuracy and uncertainty quality.

\noindent \textbf{Latent Manifold Dimensions.} The number of principal components ($K$) retained for the PCA shape manifold is tailored to each dataset's native landmark topology to capture the majority of shape variance:
\begin{itemize}
    \item \textbf{300W (68x2 shape):} $K = 60$ components.
    \item \textbf{WFLW (98x2 shape):} $K = 70$ components.
    \item \textbf{COFW (29x2 shape):} $K = 45$ components.
\end{itemize}

\section{Base Model Training and Implementation Details}
\label{app:base_model_training}

\noindent \textbf{Distilled Uncertainty Supervision Extraction for Base Model Training.} For training each base network $\Phi_{\theta}$, we construct a Test-Time Augmentation (TTA)-based uncertainty extraction wrapper around its corresponding frozen pre-trained teacher model, that produces both the point estimate $\mathbf{S}' \in \mathbb{R}^{N \times 2}$ and a per-landmark heteroscedastic uncertainty $\boldsymbol{\sigma}' \in \mathbb{R}^{N \times 2}$. The wrapper constructs an ensemble of $M = 30$ augmented views per image: for each view $i \in \{1, \ldots, M-1\}$, stochastic spatial transformations, rotation $\theta \sim \mathcal{U}(-35^{\circ}, 35^{\circ})$, isotropic scale $s \sim \mathcal{U}(1.1, 1.4)$, and translation $(t_x, t_y) \sim \mathcal{U}(-0.08, 0.08) \times (w, h)$, are applied alongside photometric perturbations (Gaussian blur with kernel $k \in \{3, 5\}$, probability 0.5) and horizontal flipping with symmetric landmark remapping (probability 0.5). The first sample ($i = 0$) is the unaugmented, deterministic prediction $\mathbf{S}' = \mathbf{S}_0$. Each transformed prediction is projected back to the original coordinate frame via the inverse affine $\mathcal{T}_i^{-1}$, and the per-landmark uncertainty is computed as:
\begin{equation}
    \boldsymbol{\sigma}' = \sqrt{\frac{1}{M} \sum_{i=0}^{M-1} (\mathbf{S}_i - \mathbf{S}')^2}
    \label{eq:sigma_tta}
\end{equation}
To prevent numerical instability, $\boldsymbol{\sigma}'$ is clipped to $[10^{-6},\; 0.5 \times d_{\text{IOD}}]$, where $d_{\text{IOD}}$ is the inter-ocular distance.

We train (a) coordinate-regression base models with heteroscedastic variance heads, and (b) heatmap-regression base models with anisotropic Gaussian targets. Since LUVLi~\citep{kumar2020luvli} and DSLPT~\citep{xia2023robust} inherently output native uncertainty estimates through their probabilistic formulations, they are utilized out-of-the-box and are not fine-tuned. For the remaining four baseline models (ORFormer~\citep{chiang2025orformer}, HR-noSA~\citep{yang2025heatmap}, PIPNet~\citep{jin2021pixel}, and STAR~\citep{zhou2023star}), we introduce uncertainty prediction capabilities tailored to their respective paradigms:

\noindent \textbf{Heatmap Regression Base Models:}
For networks that predict spatial probability maps, we replace the standard isotropic Gaussian ground-truth targets with dynamic, anisotropic Gaussian heatmaps. The spatial covariance matrix of the Gaussian target for each landmark is constructed directly from the distilled empirical uncertainty $\boldsymbol{\sigma}'$. The base model is fine-tuned to minimize the divergence between its predicted heatmaps and these uncertainty-aware Gaussian targets. Consequently, the spread of the predicted heatmaps intrinsically encodes the aleatoric uncertainty, which can be extracted at inference time.

\noindent \textbf{Coordinate Regression Base Models:}
For networks that directly regress spatial coordinates, we augment the final prediction module with a parallel heteroscedastic variance head. This auxiliary head is tasked with outputting a strictly positive variance estimate corresponding to each predicted landmark coordinate. During fine-tuning, the network minimizes a Gaussian Negative Log-Likelihood (NLL) loss, effectively forcing the variance head to match the empirical spatial covariance targets $\boldsymbol{\sigma}'$ distilled from the teacher ensemble.

\section{Evaluation Metric Definitions}
\label{app:metric_definitions}

\textbf{Normalized Mean Error (NME).} For a predicted shape $\hat{\mathbf{S}} \in \mathbb{R}^{N \times 2}$ and ground-truth $\mathbf{S}_{\text{gt}}$, the NME is:
\begin{equation}
    \text{NME} = \frac{1}{|\mathcal{I}|} \sum_{n \in \mathcal{I}} \frac{\|\hat{\mathbf{s}}_n - \mathbf{s}^{\text{gt}}_n\|_2}{d_{\text{IOD}}} \times 100\%
    \label{eq:nme}
\end{equation}
where $\mathcal{I}$ denotes the index set over which the error is computed ($\mathcal{A}$ for all, $\mathcal{O}$ for occluded).

\textbf{In-Distribution Rate (IDR).} We project the predicted shape into the PCA latent space to obtain coefficients $\mathbf{b} = \mathbf{P}^T(\hat{\mathbf{s}} - \bar{\mathbf{S}})$. A shape is classified as structurally valid (\textit{in-distribution}) if it satisfies a dual condition:
\begin{equation}
    \text{IDR}: \quad \sum_{k=1}^K b_k^2 \leq \chi^2_{K, 0.95} \quad \text{and} \quad \|\hat{\mathbf{s}} - (\bar{\mathbf{S}} + \mathbf{P}\mathbf{b})\|_2 < 0.05
    \label{eq:idr}
\end{equation}
The first condition tests whether the shape lies within the 95\% confidence region of a $\chi^2$ distribution with $K$ degrees of freedom, and the second bounds the Mean Reconstruction Error (MRE).

\textbf{Total Predictive Uncertainty.} Because \sbayes outputs a full posterior $\mathcal{N}(\mathbf{b}_\mu, \tilde{\mathbf{\Sigma}}_b)$, the total predictive uncertainty at each landmark combines epistemic and aleatoric components via the Law of Total Variance:
\begin{equation}
    \boldsymbol{\sigma}^2_{\text{pred}} = \underbrace{\text{Var}_{\mathbf{b} \sim p(\mathbf{b}|\cdot)}[\hat{\mathbf{S}}(\mathbf{b})]}_{\text{Epistemic}} + \underbrace{{\boldsymbol{\sigma}'}^2}_{\text{Aleatoric}}
    \label{eq:total_var}
\end{equation}

\section{Dynamic Occlusion: Asset Catalogue and Placement Specifications}
\label{app:occlusion_assets}

This appendix details the two complementary occlusion synthesis pipelines used to construct the occluded test sets for evaluation (Section~\ref{sec:experimental_setup}). Both pipelines operate on all three benchmarks (300W with 68-point, COFW with 29-point, and WFLW with 98-point landmark topologies), with dataset-specific anchor mappings derived from each annotation scheme.

\subsection{Pipeline 1: Wearable Face Masks (MaskTheFace)}
\label{app:masktheface}

We use MaskTheFace~\citep{anwar2020masked} to render five categories of wearable face masks: \textit{surgical}, \textit{N95}, \textit{KN95}, \textit{cloth}, and \textit{gas}. For each test image, one mask type is randomly sampled (uniform over the five categories when \texttt{mask\_type=random}). The mask fitting procedure operates as follows:

\begin{enumerate}
    \item \textbf{Landmark-to-Facial-Geometry Mapping.} Ground-truth landmarks are converted into a facial feature dictionary (jawline, nose bridge, eye contours, lip contours) using dataset-specific index mappings. For 300W, this uses the standard iBUG 68-point topology. For COFW (29 points), a synthetic 17-point jawline is constructed from the inter-pupillary axis, nose tip, and chin landmark by fitting an elliptical arc parameterized by the inter-ocular distance (IOD). For WFLW (98 points), native contour indices are used directly.
    \item \textbf{Six-Point Affine Fit.} Six anchor points on the face (derived from the nose bridge, chin, and jawline extremes) define an affine warp that maps the canonical mask template onto the subject's facial geometry, accounting for head pose (yaw, pitch, roll).
    \item \textbf{Mask Compositing.} The warped mask is alpha-composited onto the original image. A binary mask of the composited region is retained for per-landmark occlusion labelling.
\end{enumerate}

\subsection{Pipeline 2: Real-World Object Overlays}
\label{app:object_overlays}

We curate a library of 80+ RGBA object assets spanning diverse real-world occluders. Each asset undergoes alpha-channel cleaning (binary thresholding at $\alpha = 200$, followed by 2 iterations of $3 \times 3$ morphological erosion to eliminate translucent edge artefacts) and tight bounding-box cropping. The precise opaque pixel count is recorded for accurate area-based scaling.

\noindent \textbf{Anatomically-Aware Placement Modes.} Each asset is assigned to exactly one of eight placement modes, which determines (i) the facial anchor point for centering, (ii) the target face-area percentage range for dynamic scaling, and (iii) the rotation offset relative to the subject's head roll:

\begin{table}[H]
\centering
\caption{Object overlay placement modes and representative assets.}
\label{tab:placement_modes}
\small
\setlength{\tabcolsep}{4pt}
\begin{tabular}{llll}
\toprule
\textbf{Mode} & \textbf{Anchor Point} & \textbf{Face Area \%} & \textbf{Representative Assets} \\
\midrule
Global Scene & Full image & 100\% & Chainlink fence, frost, foliage, prison bars \\
Helmet & Inter-ocular midpoint & 110--140\% & Navy/silver, white/blue helmets \\
Eyewear & Inter-ocular midpoint & 15--35\% & Goggles, VR headset, masquerade mask \\
Faceguard & Nose--chin midpoint & 25--40\% & Football helmet faceguard \\
Mouth & Mouth centroid & 5--40\% & Mug, burger, pizza, microphones, fruit \\
Camera & Random eye/midpoint & 20--40\% & DSLR, telephoto lens, TLR camera \\
Hand/Glove & Random facial feature & 20--45\% & Boxing glove, fists, open palm, mittens \\
Lower-Mid Blocker & Nose--mouth midpoint & 15--60\% & Tablet, teddy, bouquet, shipping box \\
Floating & Random landmark & 5--50\% & Football, soccer ball, balloons, palm frond \\
\bottomrule
\end{tabular}
\end{table}

\noindent \textbf{Dynamic Area Scaling.} For each asset, a target face-area percentage $p$ is uniformly sampled from the mode-specific range. The face area is estimated as an ellipse $A_{\text{face}} = \pi \cdot (w_{\text{face}}/2) \cdot (h_{\text{face}}/2)$, where $w_{\text{face}}$ is the outer face width and $h_{\text{face}} = 1.25 \times \|\text{eye\_mid} - \text{chin}\|_2$. The asset is then resized by scale factor $\sqrt{p \cdot A_{\text{face}} / A_{\text{opaque}}}$, where $A_{\text{opaque}}$ is the pre-computed opaque pixel count, ensuring the \textit{actual occluding area} (not bounding box) matches the target.

\noindent \textbf{Occlusion Label Extraction.} After compositing, the alpha channel of the overlay defines a continuous occlusion mask $\alpha(x, y) \in [0, 1]$. A landmark $n$ is classified as occluded if $\alpha(\mathbf{s}_n^{\text{gt}}) > 0.5$, producing the per-image partition $(\mathcal{A}, \mathcal{O})$ used for decoupled evaluation.

\section{Data-Driven Prior: Training Augmentation Protocol}
\label{app:training_augmentation}

To train the \sbayes Transformer encoder $E_\phi$, we construct a diverse dataset of realistic $(\mathbf{S}_{\text{gt}}, \mathbf{S}', \boldsymbol{\sigma}')$ triplets by procedurally corrupting each pristine training image with 13 structured augmentation variants. All spatial parameters are anchored to the Inter-Ocular Distance for scale invariance (e.g., $d_{\text{IOD}} = \|\mathbf{s}_{36} - \mathbf{s}_{45}\|_2$ for 300W). The ratio of non-occluding to occluding variants (4:9) ensures the solver observes confident predictions (low $\boldsymbol{\sigma}'$) approximately 30\% of the time, preventing the prior from being unconditionally strong.

\subsection{Non-Occluding Photometric Variants}

These variants degrade image quality without directly obscuring facial landmarks, producing training pairs where the base model remains largely confident ($\boldsymbol{\sigma}' \approx 0$):

\begin{enumerate}
    \item \textbf{Pristine:} The unmodified training image.
    \item \textbf{Brightness (Gamma Correction):} $\gamma \sim \mathcal{U}(0.55, 0.8)$. Applies a power-law LUT: $I_{\text{out}}(p) = \lfloor (I_{\text{in}}(p) / 255)^\gamma \times 255 \rceil$.
    \item \textbf{Contrast (Linear Adjustment):} Gain $\alpha \sim \mathcal{U}(0.5, 1.8)$, bias $\beta \sim \mathcal{U}(-30, 30)$. Applies $I_{\text{out}} = \text{clip}(\alpha \cdot I_{\text{in}} + \beta, 0, 255)$.
    \item \textbf{Color Shift (HSV Perturbation):} Hue offset $\Delta h \sim \mathcal{U}(-15, 15)$ (mod 180), saturation scale $\alpha_s \sim \mathcal{U}(0.6, 1.4)$.
\end{enumerate}

\subsection{Occluding Structural Variants}

These variants introduce spatially structured occlusions that selectively corrupt subsets of landmarks:

\begin{enumerate}
    \item \textbf{Lower-Half Mask:} A grey rectangle ($128, 128, 128$) with additive noise ($\mathcal{U}(0, 50)$) is placed from the nose tip (landmark 33) $\pm \mathcal{U}(-0.3, 0.3) \times d_{\text{IOD}}$ downward to the image bottom.

    \item \textbf{Anatomical Sunglasses/Eyepatches:} Dark rectangles ($\text{RGB} \sim \mathcal{U}(5, 30)^3$) are fitted to bounding boxes of the eye and eyebrow landmarks (36--47) with margin $\mathcal{U}(0.2, 0.6) \times d_{\text{IOD}}$. Three asymmetric variants are sampled: both eyes ($p = 0.6$), left only ($p = 0.2$), or right only ($p = 0.2$). When both eyes are occluded, a connecting bridge strip spans the nose bridge (landmarks 27--28).

    \item \textbf{Spatially Variant Blur:} A Gaussian blur with kernel size $k = \lfloor 0.8 \times d_{\text{IOD}} \rceil | 1$ is applied through a spatially varying blend mask generated from low-frequency random noise ($h/8 \times w/8$, bicubic upsampled, Gaussian smoothed, scaled $\times 1.5$ and clipped to $[0, 1]$).

    \item \textbf{Extreme Random Cutout:} A rectangular patch of size $\mathcal{U}(0.5, 1.5) \times d_{\text{IOD}}$ (both axes) is placed at a random image location and filled with uniform random pixel noise $\sim \mathcal{U}(0, 255)^3$.

    \item \textbf{Random Lines:} $n \sim \mathcal{U}(5, 15)$ line segments with random endpoints, random RGB color, and thickness $\max(1, \lfloor d_{\text{IOD}} \times \mathcal{U}(0.01, 0.05) \rceil)$ are drawn across the image, simulating stray hair, wires, or scratches.

    \item \textbf{Vertical Half-Face Block:} A solid-color rectangle is placed from the nose bridge (landmark 30) $\pm \mathcal{U}(-0.1, 0.1) \times d_{\text{IOD}}$ to either the left or right image boundary (randomly selected with equal probability), simulating wall edges or doorframes.

    \item \textbf{Additive Sensor Noise:} Gaussian noise $\eta \sim \mathcal{N}(0, \sigma_n^2)$ with $\sigma_n \sim \mathcal{U}(10, 30)$ is added pixel-wise, simulating low-light sensor artifacts.

    \item \textbf{Irregular Elliptical Blobs:} $n \sim \mathcal{U}(1, 6)$ filled ellipses with semi-axes $\mathcal{U}(0.2, 0.6) \times d_{\text{IOD}}$, random rotation, and skin-like color ($\text{RGB} \sim \mathcal{U}(180, 250)^3$) are anchored to jaw and mouth landmarks (indices 0--16 and 48--67) with positional jitter $\pm 0.5 \times d_{\text{IOD}}$, simulating hands or objects partially covering the lower face.

    \item \textbf{Targeted Micro-Occlusions:} $n \sim \mathcal{U}(2, 6)$ randomly selected individual landmarks receive small circular or rectangular occlusions with radius $\mathcal{U}(0.08, 0.2) \times d_{\text{IOD}}$ and random RGB color, simulating localized debris, birthmarks, or sensor defects.
\end{enumerate}

\noindent For each of the 13 variants, the frozen base model $\Phi_T$ is queried with $M = 30$ TTA passes (Section~\ref{sec:method}) to produce empirically grounded predictions $\mathbf{S}'$ and uncertainties $\boldsymbol{\sigma}'$. The per-landmark $\boldsymbol{\sigma}'$ is clipped to $[10^{-6},\; 0.5 \times d_{\text{IOD}}]$. The resulting $13 \times |\mathcal{D}_{\text{train}}|$ triplets are serialized and used for offline \sbayes Transformer training.

\section{Diverse Posterior Sampling for minNME}
\label{app:diverse_sampling}

The minNME metric (Section~\ref{sec:experimental_setup}) requires drawing samples from the posterior $\mathbf{b} \sim \mathcal{N}(\mathbf{b}_\mu, \tilde{\mathbf{\Sigma}}_b)$ that systematically cover the distribution's principal modes rather than clustering near the mean. We employ a deterministic eigenvector-based sampling scheme:

\begin{enumerate}
    \item Compute the eigendecomposition $\tilde{\mathbf{\Sigma}}_b = \mathbf{V} \boldsymbol{\Lambda} \mathbf{V}^T$, where $\boldsymbol{\Lambda} = \text{diag}(\lambda_1, \ldots, \lambda_K)$ are eigenvalues sorted in decreasing order and $\mathbf{V} = [\mathbf{v}_1, \ldots, \mathbf{v}_K]$ are the corresponding eigenvectors.
    \item For each principal component $k \in \{1, \ldots, K\}$, generate two samples:
    \begin{equation}
        \mathbf{b}_{2k-1} = \mathbf{b}_\mu + 2\sqrt{\lambda_k}\,\mathbf{v}_k, \qquad \mathbf{b}_{2k} = \mathbf{b}_\mu - 2\sqrt{\lambda_k}\,\mathbf{v}_k
    \end{equation}
    which places samples at $\pm 2\sigma$ along each eigenvector direction.
    \item If $N_s > 2K$, additional samples are drawn via standard multivariate normal sampling.
\end{enumerate}

This scheme ensures that the first $2K$ samples are located at the 95th percentile boundary along each independent axis of variation, maximizing the chance of finding a sample near the ground truth even when the posterior mean is displaced by structured error.

\FloatBarrier



%% file: references.bib
@article{feldman2006bayesian,
  title={{Bayesian} estimation of the shape skeleton},
  author={Feldman, Jacob and Singh, Manish},
  journal={Proceedings of the National Academy of Sciences},
  volume={103},
  number={47},
  pages={18014--18019},
  year={2006},
  publisher={National Academy of Sciences}
}

@inproceedings{xu2024bayesian,
  title={{Bayesian} diffusion models for {3D} shape reconstruction},
  author={Xu, Haiyang and Lei, Yu and Chen, Zeyuan and Zhang, Xiang and Zhao, Yue and Wang, Yilin and Tu, Zhuowen},
  booktitle={2024 IEEE/CVF Conference on Computer Vision and Pattern Recognition (CVPR)},
  pages={10628--10638},
  year={2024},
  organization={IEEE}
}

@article{kersten2004object,
  title={Object perception as {Bayesian} inference},
  author={Kersten, Daniel and Mamassian, Pascal and Yuille, Alan},
  journal={Annu. Rev. Psychol.},
  volume={55},
  number={1},
  pages={271--304},
  year={2004},
  publisher={Annual Reviews}
}

@article{mamassian2006bayesian,
  title={{Bayesian} inference of form and shape},
  author={Mamassian, Pascal},
  journal={Progress in Brain Research},
  volume={154},
  pages={265--270},
  year={2006},
  publisher={Elsevier}
}

@inproceedings{mio2004learning,
  title={Learning and {Bayesian} shape extraction for object recognition},
  author={Mio, Washington and Srivastava, Anuj and Liu, Xiuwen},
  booktitle={European Conference on Computer Vision},
  pages={62--73},
  year={2004},
  organization={Springer}
}

@inproceedings{wang2026pose,
  title={Pose prior learner: Unsupervised categorical prior learning for pose estimation},
  author={Wang, Ziyu and Han, Shuangpeng and Zhang, Mengmi},
  booktitle={International Conference on Learning Representations},
  volume={2026},
  pages={141125--141146},
  year={2026}
}

@inproceedings{zheng2025hipart,
  title={{HiPART}: Hierarchical pose autoregressive transformer for occluded {3D} human pose estimation},
  author={Zheng, Hongwei and Li, Han and Dai, Wenrui and Zheng, Ziyang and Li, Chenglin and Zou, Junni and Xiong, Hongkai},
  booktitle={2025 IEEE/CVF Conference on Computer Vision and Pattern Recognition (CVPR)},
  pages={16807--16817},
  year={2025},
  organization={IEEE}
}

@article{zheng2026pandapose,
  title={{PandaPose}: {3D} Human Pose Lifting from a Single Image via Propagating {2D} Pose Prior to {3D} Anchor Space},
  author={Zheng, Jinghong and Jiang, Changlong and Xiao, Yang and Li, Jiaqi and Kuang, Haohong and Xu, Hang and Wang, Ran and Cao, Zhiguo and Du, Min and Zhou, Joey Tianyi},
  journal={Advances in Neural Information Processing Systems},
  volume={38},
  pages={90147--90169},
  year={2026}
}

@inproceedings{yang2025heatmap,
  title={Heatmap Regression without Soft-Argmax for Facial Landmark Detection},
  author={Yang, Chiao-An and Yeh, Raymond A},
  booktitle={2025 IEEE/CVF International Conference on Computer Vision (ICCV)},
  pages={28729--28739},
  year={2025},
  organization={IEEE}
}

@inproceedings{wang2019adaptive,
  title={Adaptive wing loss for robust face alignment via heatmap regression},
  author={Wang, Xinyao and Bo, Liefeng and Fuxin, Li},
  booktitle={2019 IEEE/CVF International Conference on Computer Vision (ICCV)},
  pages={6970--6980},
  year={2019},
  organization={IEEE}
}

@article{xu2022vitpose,
  title={{ViTPose}: Simple vision transformer baselines for human pose estimation},
  author={Xu, Yufei and Zhang, Jing and Zhang, Qiming and Tao, Dacheng},
  journal={Advances in Neural Information Processing Systems},
  volume={35},
  pages={38571--38584},
  year={2022}
}

@inproceedings{zhou2023star,
  title={{STAR} loss: Reducing semantic ambiguity in facial landmark detection},
  author={Zhou, Zhenglin and Li, Huaxia and Liu, Hong and Wang, Nanyang and Yu, Gang and Ji, Rongrong},
  booktitle={2023 IEEE/CVF Conference on Computer Vision and Pattern Recognition (CVPR)},
  pages={15475--15484},
  year={2023},
  organization={IEEE}
}

@article{jin2021pixel,
  title={Pixel-in-pixel net: Towards efficient facial landmark detection in the wild},
  author={Jin, Haibo and Liao, Shengcai and Shao, Ling},
  journal={International Journal of Computer Vision},
  volume={129},
  number={12},
  pages={3174--3194},
  year={2021},
  publisher={Springer}
}

@article{xia2023robust,
  title={Robust face alignment via inherent relation learning and uncertainty estimation},
  author={Xia, Jiahao and Xu, Min and Zhang, Haimin and Zhang, Jianguo and Huang, Wenjian and Cao, Hu and Wen, Shiping},
  journal={IEEE Transactions on Pattern Analysis and Machine Intelligence},
  volume={45},
  number={8},
  pages={10358--10375},
  year={2023},
  publisher={IEEE}
}

@inproceedings{chiang2025orformer,
  title={{ORFormer}: Occlusion-robust transformer for accurate facial landmark detection},
  author={Chiang, Jui-Che and Hu, Hou-Ning and Hou, Bo-Syuan and Tseng, Chia-Yu and Liu, Yu-Lun and Chen, Min-Hung and Lin, Yen-Yu},
  booktitle={2025 IEEE/CVF Winter Conference on Applications of Computer Vision (WACV)},
  pages={784--793},
  year={2025},
  organization={IEEE}
}

@inproceedings{kumar2020luvli,
  title={{LUVLi} face alignment: Estimating landmarks' location, uncertainty, and visibility likelihood},
  author={Kumar, Abhinav and Marks, Tim K and Mou, Wenxuan and Wang, Ye and Jones, Michael and Cherian, Anoop and Koike-Akino, Toshiaki and Liu, Xiaoming and Feng, Chen},
  booktitle={2020 IEEE/CVF Conference on Computer Vision and Pattern Recognition (CVPR)},
  pages={8233--8243},
  year={2020},
  organization={IEEE}
}

@article{tian2025test,
  title={Test-time Uncertainty Estimation for Medical Image Registration via Transformation Equivariance},
  author={Tian, Lin and Hu, Xiaoling and Eugenio Iglesias, Juan},
  journal={arXiv e-prints},
  pages={arXiv--2509},
  year={2025}
}

@inproceedings{xiang2026occface,
  title={{OccFace}: Unified Occlusion-Aware Facial Landmark Detection with Per-Point Visibility},
  author={Xiang, Xinhao and Li, Zhengxin and Dhakad, Saurav and Bancroft, Theo and Zhang, Jiawei and Li, Weiyang},
  booktitle={International Conference on Pattern Recognition},
  pages={202--217},
  year={2026},
  organization={Springer}
}

@article{li2024cascaded,
  title={Cascaded iterative transformer for jointly predicting facial landmark, occlusion probability and head pose},
  author={Li, Yaokun and Tan, Guang and Gou, Chao},
  journal={International Journal of Computer Vision},
  volume={132},
  number={4},
  pages={1242--1257},
  year={2024},
  publisher={Springer}
}

@inproceedings{zhu2022occlusion,
  title={Occlusion-robust face alignment using a viewpoint-invariant hierarchical network architecture},
  author={Zhu, Congcong and Wan, Xintong and Xie, Shaorong and Li, Xiaoqiang and Gu, Yinzheng},
  booktitle={2022 IEEE/CVF Conference on Computer Vision and Pattern Recognition (CVPR)},
  pages={11102--11111},
  year={2022},
  organization={IEEE}
}

@article{zhu2022reasoning,
  title={Reasoning structural relation for occlusion-robust facial landmark localization},
  author={Zhu, Congcong and Li, Xiaoqiang and Li, Jide and Dai, Songmin and Tong, Weiqin},
  journal={Pattern Recognition},
  volume={122},
  pages={108325},
  year={2022},
  publisher={Elsevier}
}

@article{cootes1995active,
  title={Active shape models---their training and application},
  author={Cootes, Timothy F and Taylor, Christopher J and Cooper, David H and Graham, Jim},
  journal={Computer Vision and Image Understanding},
  volume={61},
  number={1},
  pages={38--59},
  year={1995},
  publisher={Elsevier}
}

@article{anwar2020masked,
  title={Masked Face Recognition for Secure Authentication},
  author={Anwar, Aqeel and Raychowdhury, Arijit},
  journal={arXiv preprint arXiv:2008.11104},
  year={2020}
}

@inproceedings{cofw,
  title={Robust face landmark estimation under occlusion},
  author={Burgos-Artizzu, Xavier P. and Perona, Pietro and Doll{\'a}r, Piotr},
  booktitle={Proceedings of the IEEE International Conference on Computer Vision (ICCV)},
  pages={1513--1520},
  year={2013}
}

@inproceedings{300w,
  title={300 faces in-the-wild challenge: The first facial landmark localization challenge},
  author={Sagonas, Christos and Tzimiropoulos, Georgios and Zafeiriou, Stefanos and Pantic, Maja},
  booktitle={Proceedings of the IEEE International Conference on Computer Vision Workshops},
  pages={397--403},
  year={2013}
}

@inproceedings{wflw,
  title={Look at boundary: A boundary-aware face alignment algorithm},
  author={Wu, Wayne and Qian, Chen and Yang, Shuo and Wang, Quan and Cai, Yici and Zhou, Qiang},
  booktitle={Proceedings of the IEEE Conference on Computer Vision and Pattern Recognition (CVPR)},
  pages={2129--2138},
  year={2018}
}

@article{lee2018simple,
  title={A simple unified framework for detecting out-of-distribution samples and adversarial attacks},
  author={Lee, Kimin and Lee, Kibok and Lee, Honglak and Shin, Jinwoo},
  journal={Advances in Neural Information Processing Systems},
  volume={31},
  year={2018}
}

@inproceedings{kingma2013auto,
  author    = {Diederik P. Kingma and Max Welling},
  title     = {Auto-Encoding Variational {Bayes}},
  booktitle = {International Conference on Learning Representations (ICLR)},
  year      = {2014}
}

@inproceedings{li2021human,
  title={Human Pose Regression with Residual Log-likelihood Estimation},
  author={Li, Jiefeng and Bian, Can and Zeng, Ailing and Wang, Can and Pang, Bo and Liu, Wentao and Li, Cewu},
  booktitle={Proceedings of the IEEE/CVF International Conference on Computer Vision (ICCV)},
  pages={11025--11034},
  year={2021}
}

@inproceedings{neven2021dirichlet,
  title={Dirichlet Regressive Mapping for Keypoint Localization},
  author={Neven, Davy and De Brabandere, Bert and Proesmans, Marc and Van Gool, Luc},
  booktitle={Proceedings of the IEEE/CVF Conference on Computer Vision and Pattern Recognition (CVPR)},
  pages={12467--12476},
  year={2021}
}

@article{jonkers2025reliable,
  title={Reliable uncertainty quantification for {2D/3D} anatomical landmark localization using multi-output conformal prediction},
  author={Jonkers, Jef and Coopman, Frank and Duchateau, Luc and Van Wallendael, Glenn and Van Hoecke, Sofie},
  journal={arXiv preprint arXiv:2503.14106},
  year={2025}
}
